\documentclass{article}

\PassOptionsToPackage{table}{xcolor}

\usepackage[main, final]{neurips_2026}

\usepackage[utf8]{inputenc}
\usepackage[T1]{fontenc}
\usepackage{hyperref}
\usepackage{url}
\usepackage{booktabs}
\usepackage{amsfonts}
\usepackage{nicefrac}
\usepackage{microtype}
\usepackage{xcolor}
\usepackage{graphicx}
\usepackage{subcaption}
\usepackage{multirow}
\usepackage{enumitem}
\usepackage{amsmath}
\usepackage{amssymb}
\usepackage{mathtools}
\usepackage{amsthm}
\usepackage[capitalize,noabbrev]{cleveref}
\usepackage[table]{xcolor}
\usepackage{caption}
\usepackage{wrapfig}

\usepackage{amsmath,amsfonts,bm}

\def\eqref#1{equation~\ref{#1}}

\def\1{\bm{1}}

\def\vx{{\bm{x}}}

\DeclareMathAlphabet{\mathsfit}{\encodingdefault}{\sfdefault}{m}{sl}
\SetMathAlphabet{\mathsfit}{bold}{\encodingdefault}{\sfdefault}{bx}{n}

\theoremstyle{plain}

\theoremstyle{definition}

\theoremstyle{remark}

\title{UltraDiff: Transferring High-Fidelity Priors to Compressed Latent Spaces for High-Resolution Image Generation}

\author{%
  \normalfont\mdseries
 \textbf{Jingjing Ren}\textsuperscript{1,*}\quad
  \textbf{Haitian Zheng}\textsuperscript{2,*}\quad
  \textbf{Haoyu Chen}\textsuperscript{1}\quad
  \textbf{Connelly Barnes}\textsuperscript{2}\\
  \textbf{Zhifei Zhang}\textsuperscript{2}\quad
  \textbf{Yuqian Zhou}\textsuperscript{2}\quad
  \textbf{Zhe Lin}\textsuperscript{2}\quad
  \textbf{Lei Zhu}\textsuperscript{1,\textdagger}\\[0.4em]
  \textsuperscript{1}The Hong Kong University of Science and Technology (Guangzhou)\\
  \textsuperscript{2}Adobe Research\\[0.25em]
  \textsuperscript{*}Equal contribution
  \qquad
  \textsuperscript{\textdagger}Corresponding author
}

\begin{document}

\maketitle

\begin{abstract}
High-resolution synthesis with Latent Diffusion Models (LDMs) faces a fundamental trade-off between computational cost and generative fidelity.
This trade-off is largely governed by the latent compression factor $f$ (e.g. f8: $f{=}8$ and f32: $f{=}32$).
While aggressive compression (f32) significantly improves efficiency, it suffers from severe information loss and ambiguity compared to standard f8 models.
We propose UltraDiff, a framework that reformulates this challenge as joint-latent diffusion learning followed by marginal score distillation.
We introduce MixTeacher, a dual-branch DiT that models the joint f8-f32 distribution, utilizing shared attention to resolve the ambiguity of the compressed space with high-fidelity f8 priors.
Crucially, we propose vector field distillation in the native f32 space. Theoretically, this compels the student to learn the conditional expectation of the teacher's trajectory, effectively filtering out compression artifacts.
Across the evaluated 1K--4K settings, UltraDiff approaches f8-level quality while preserving f32 efficiency. With 20 sampling steps on a single A100 GPU, it generates a 2K image in 6.2 seconds---over $40\times$ faster than our f8 baseline.
\end{abstract}

\section{Introduction}
\label{sec:intro}
\begin{figure*}[!h]
  \centering
   \includegraphics[width=0.9\linewidth]{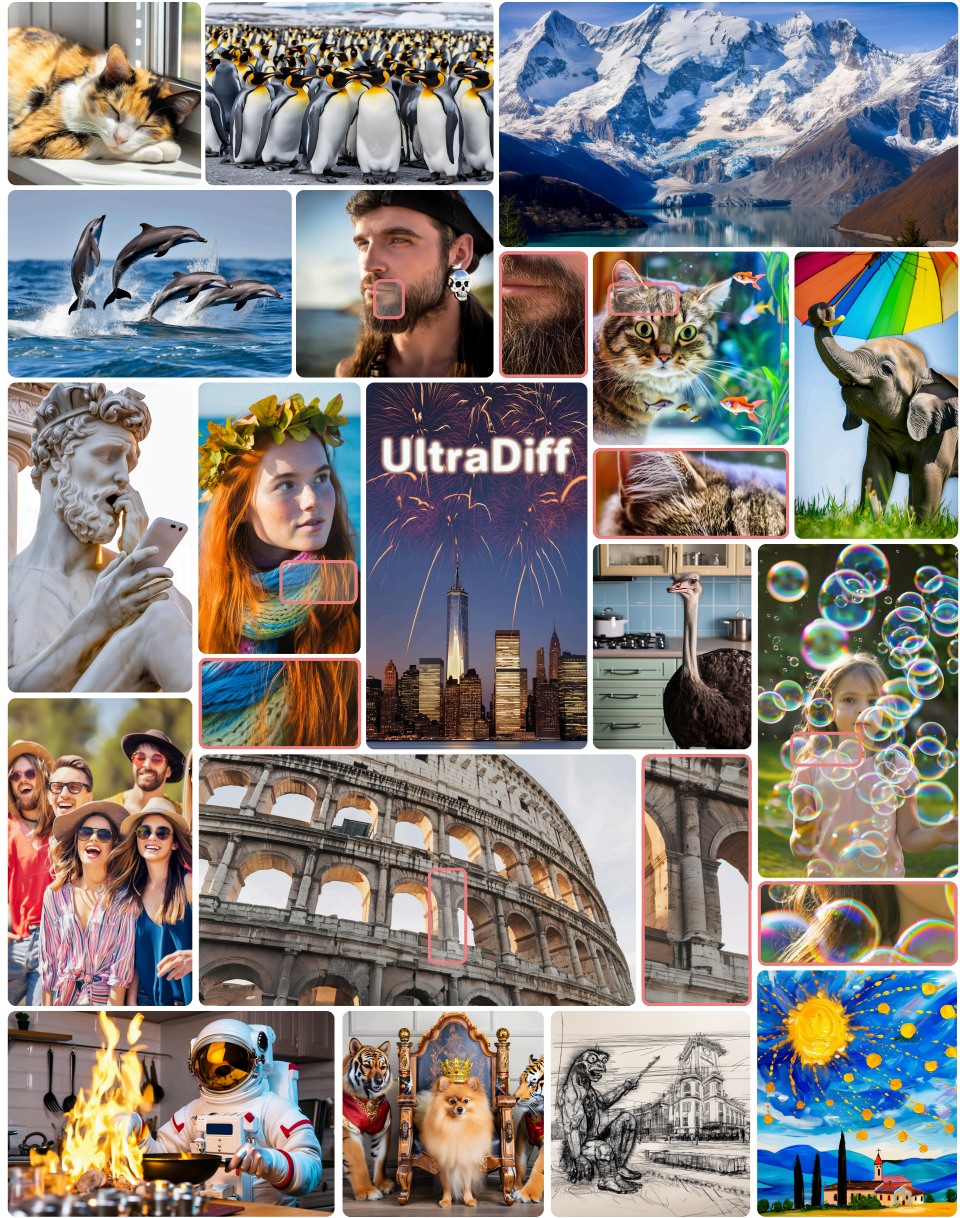}
   \caption{Our UltraDiff generates high-quality images across a variety of scenes, showcasing rich details and diverse content. Please zoom in to view the details. Additional images are provided in the supplementary materials.
   }
   \label{fig:teaser}
\end{figure*}

The demand for ultra-high-definition (UHD) image and video synthesis has surged across streaming, gaming, and VR applications.
While recent Diffusion Transformers (DiTs)~\citep{esser2024scaling, chen2023pixart, podell2023sdxl, FLUX} excel at high-fidelity generation, self-attention dominates their compute and scales quadratically with the token count ($O(N^2)$), making UHD synthesis prohibitively expensive.
To mitigate this, recent works~\citep{hunyuanimage21_github, xie2024sana} explore VAEs with more aggressive spatial downsampling, i.e., increasing the latent downsampling factor from f8 ($f{=}8$) to f32 ($f{=}32$), reducing the 2D latent token count by $16\times$ compared to a standard f8 VAE, offering substantial efficiency gains.

However, the efficiency of aggressive downsampling comes at the cost of a substantially harder diffusion problem~\citep{skorokhodov2025improving}.
With f32 compression, the diffusion model must synthesize complex scenes from far fewer spatial tokens, yielding an ambiguous target distribution for the diffusion model to learn. This makes it difficult for the model to capture fine-grained details and precise spatial layout, leading to degraded perceptual fidelity and weaker prompt adherence compared to f8-based models~\citep{FLUX, li2024playground, kong2024hunyuanvideo}.

\begin{wrapfigure}{r}{0.5\linewidth}
  \vspace{-10pt}
  \centering
  \includegraphics[width=\linewidth]{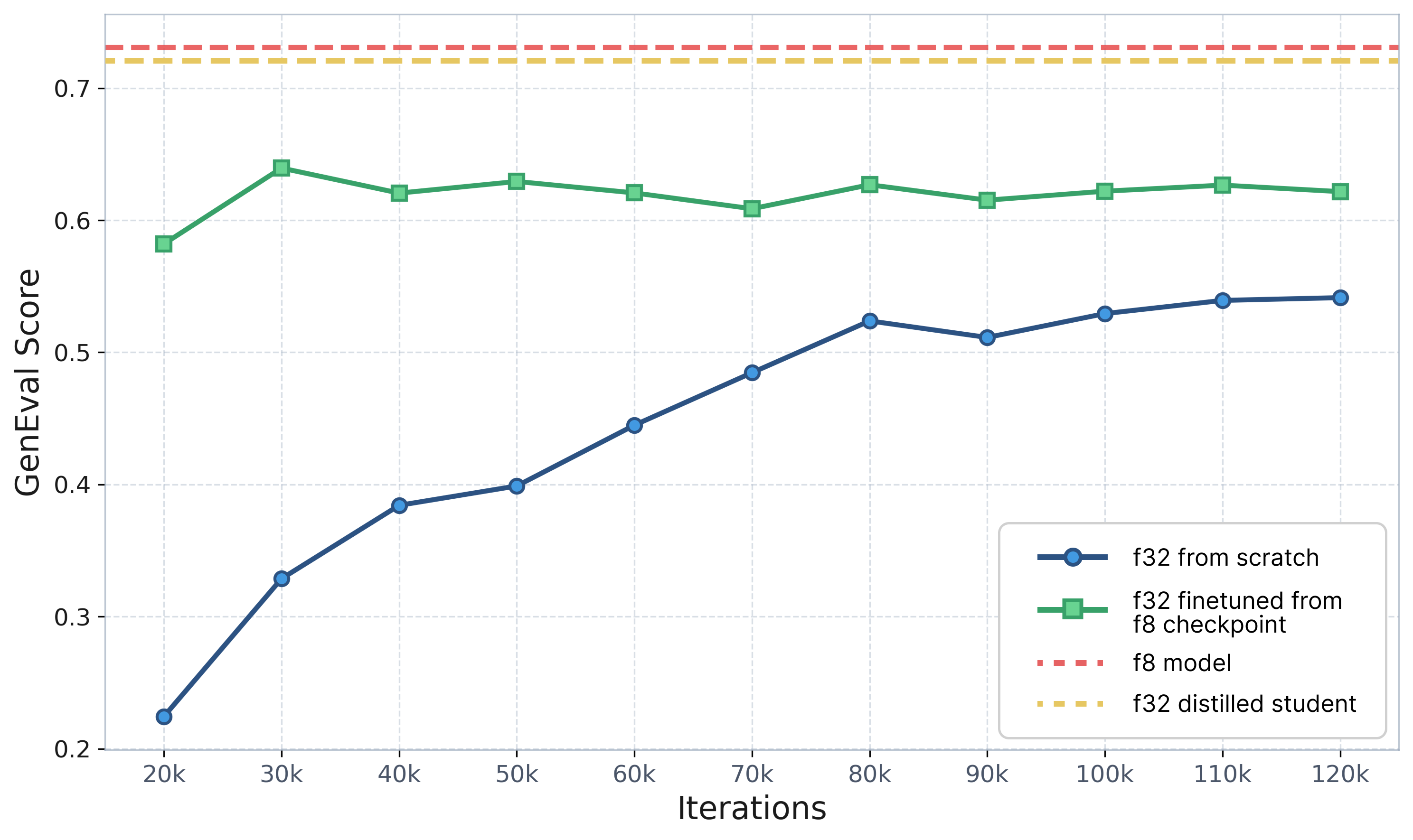}
  \caption{\textbf{Effectiveness of Cross-Space Knowledge Transfer.} GenEval scores during training. UltraDiff bridges the gap to the f8 teacher.}
  \label{fig:geneval_iter}
  \vspace{-10pt}
\end{wrapfigure}

Given the superior fidelity and detail generation capacity of f8 models, a natural question arises:
\textbf{Can we transfer the rich generative knowledge of a pre-trained f8 model to guide the training of an efficient f32 model?}
To study this, we conduct a controlled comparison between an f32 DiT trained from scratch and one initialized from a pre-trained f8 model.
As shown in Fig.~\ref{fig:geneval_iter}, f8 initialization consistently improves performance, indicating that generative knowledge indeed transfers across diffusion model with two latent spaces, despite the mismatch in latent representation, dimensionality and resolution.
However, from Fig.~\ref{fig:geneval_iter}, the benefit of initialization quickly saturates, leaving a substantial performance gap between the two models. This suggests that static parameter initialization alone is insufficient; and effective training requires dynamic, state-dependent guidance.
Instead, a more natural solution is generative knowledge distillation where most existing methods~\citep{leng2025repa, ren2025turbo2k} rely on feature alignment, forcing a student to match intermediate representations of a trained teacher model.
However, we argue that \textit{static feature alignment} are suboptimal when distilling across mismatched latent resolutions, as aligning an f32 student to an f8 teacher requires interpolating feature grids, leading to quantization artifacts and checkerboard patterns.
Moreover, such implicit feature-level regularization does not provide explicit distillation supervision, which limits their ability to transfer generative knowledge from a pre-trained diffusion teacher.

To address these challenges for efficient UHD synthesis, we propose UltraDiff, a new framework that learns diffusion in a joint f8--f32 latent space and then distills the resulting score to the f32 latent space, yielding the final diffusion model in the f32 marginal for efficient sampling.
Specifically, instead of treating the diffusion modeling in f8 and f32 spaces as two separate tasks, we introduce \textbf{MixTeacher}, a dual-branch DiT that models the joint distribution $p(\vx^{\text{f8}}, \vx^{\text{f32}})$ where $\vx^{\text{f8}}$ and $\vx^{\text{f32}}$ are extracted from the same image.
Architecturally, MixTeacher comprises two DiT branches with separate parameters to accommodate the distinct feature distributions in the f8 and f32 latent spaces, while sharing self-attention layers within each block to enable dense cross-space interaction.
With f8 initialization and layer-wise attention sharing, MixTeacher effectively learns a joint denoising model over paired f8 and f32 latents, effectively transferring generative knowledge from the pre-trained f8 branch to the f32 branch.
Crucially, unlike other implicit feature distillation methods~\citep{leng2025repa, ren2025turbo2k}, it enables direct distilling supervision in the f32 space.
Specifically, our teacher projects its high-fidelity guidance into the student's f32 space.
Theoretically, this forces the student to learn the marginal vector field, effectively approximating the conditional expectation of the high-fidelity trajectory.
This process ``integrates out'' the ambiguity inherent to compressed generation, guiding the student toward structurally coherent manifolds while operating strictly within the efficient f32 latent space.

Our contributions are summarized as follows:
\begin{itemize}
    \item We introduce MixTeacher, a unified architecture that learns the joint score function of f8 and f32 latents. Through a stochastic routing strategy and shared attention mechanism, it captures cross-space correlations while maintaining the marginal generation capabilities of each branch.
    \item We develop a marginal score distillation strategy that aligns the vector field directly in the native f32 space. By learning the conditional expectation of the teacher's trajectory, the student model effectively resolves the inherent ambiguity and difficulty of modeling complex dependencies in the compressed f32 space.
    \item Our method, UltraDiff, achieves a strong trade-off between quality and efficiency. It produces detail-rich 4K images with strong semantic alignment shown in Figure~\ref{fig:teaser}, approaches the fidelity of the evaluated f8 systems, and generates a 2K image in 6.2 seconds using 20 sampling steps on a single NVIDIA A100 GPU---$40.3\times$ faster than our f8 anchor. UltraDiff also scales to 4K generation, where the f8 anchor runs out of memory.
\end{itemize}


\vspace{-10pt}
\section{Related Work}
\label{sec:related_work}
\subsection{Diffusion Transformer}
Diffusion Transformers (DiTs) have become a widely used backbone for image and video synthesis~\citep{chen2023pixart,chen2024pixart,chen2025sana, flux2,hunyuanimage21_github,li2024playground, li2026hp, xiao2026joyai,flowar} because of their scaling behavior and strong generative performance~\citep{peebles2023scalable}. A typical system pairs a DiT with a variational autoencoder (VAE), which maps an image of size $3\times H\times W$ to a latent tensor of size $c\times H/f\times W/f$. Self-attention then operates on the latent tokens, with computational and memory costs that grow quadratically with the sequence length~\citep{chen2023pixart,ren2025turbo2k,esser2024scaling,hunyuanimage21_github,ren2025beyond,ren2026decoupled,tian2024visual}.

Most systems use an $8\times$ spatial compression factor and often reduce the token count further with $2\times2$ patch embedding~\citep{podell2023sdxl,chen2023pixart,chen2024pixart,li2024playground,esser2024scaling,FLUX}. More aggressive compression, such as $f=32$, improves efficiency for high-resolution and long-sequence generation~\citep{xie2024sana,hacohen2024ltx,hunyuanimage21_github}. However, the smaller spatial grid makes structure and semantic alignment harder to learn~\citep{xie2024sana,hacohen2024ltx,pernias2023wurstchen}. UltraDiff addresses this problem by transferring knowledge from a strong f8 model to an f32 model through initialization and cross-latent distillation, followed by mixed-resolution training.

Recent work has also explored making one autoencoder latent space explicitly compatible with another during VAE training~\citep{yi2025finestructure,zhao2024cvvae}. UltraDiff addresses a complementary setting: our f8 and f32 VAEs are trained independently and then kept fixed, while their relationship is learned at the diffusion vector-field level through shared attention in MixTeacher. Explicit compatible-VAE training may further simplify cross-space transfer and is an interesting direction for future work.

\subsection{Diffusion Distillation }
Distillation has become a general mechanism for adapting diffusion models to diverse objectives and deployment constraints. For acceleration, progressive and distribution--matching distillation compress long sampling trajectories into few- or even one-step generators while maintaining competitive fidelity \citep{salimans2022progressive, yin2024improved, yin2024one, kang2024distilling}. Related consistency and rectified-flow formulations further enable fast sampling and have been applied in high-resolution restoration pipelines \citep{xu2025fast, sun2025pocketsr, wang2024sinsr}. 
Beyond speed, distillation can change the inference pattern or model architecture. For example, a bidirectional teacher can supervise a causal student for streaming video generation~\citep{yin2024slow, huang2025self, novack2024presto, shin2025motionstream, kodaira2025streamdit}.
Distillation also supports modality transfer and prior reuse; score-distillation techniques leverage strong 2D text-to-image priors to supervise 3D synthesis, with recent work addressing stability and mode collapse \citep{yu2023text, ma2024scaledreamer, wang2024taming, wang2025videoscene}. 
Another class of methods aligns semantic features from external encoders, such as DINOv2 or other diffusion networks, with a model's internal representations to accelerate training and improve image fidelity~\citep{yu2024representation, leng2025repa, ren2025turbo2k}.
We adopt the spirit of distillation to bridge a latent-scale mismatch: a strong f8 teacher guides an f32 DiT via initialization and cross-latent supervision, yielding a high-compression student that retains fidelity while benefiting from the computational savings of far fewer tokens.



\section{Method}
\begin{figure}[htbp!]
\centering\includegraphics[width=0.85\linewidth]{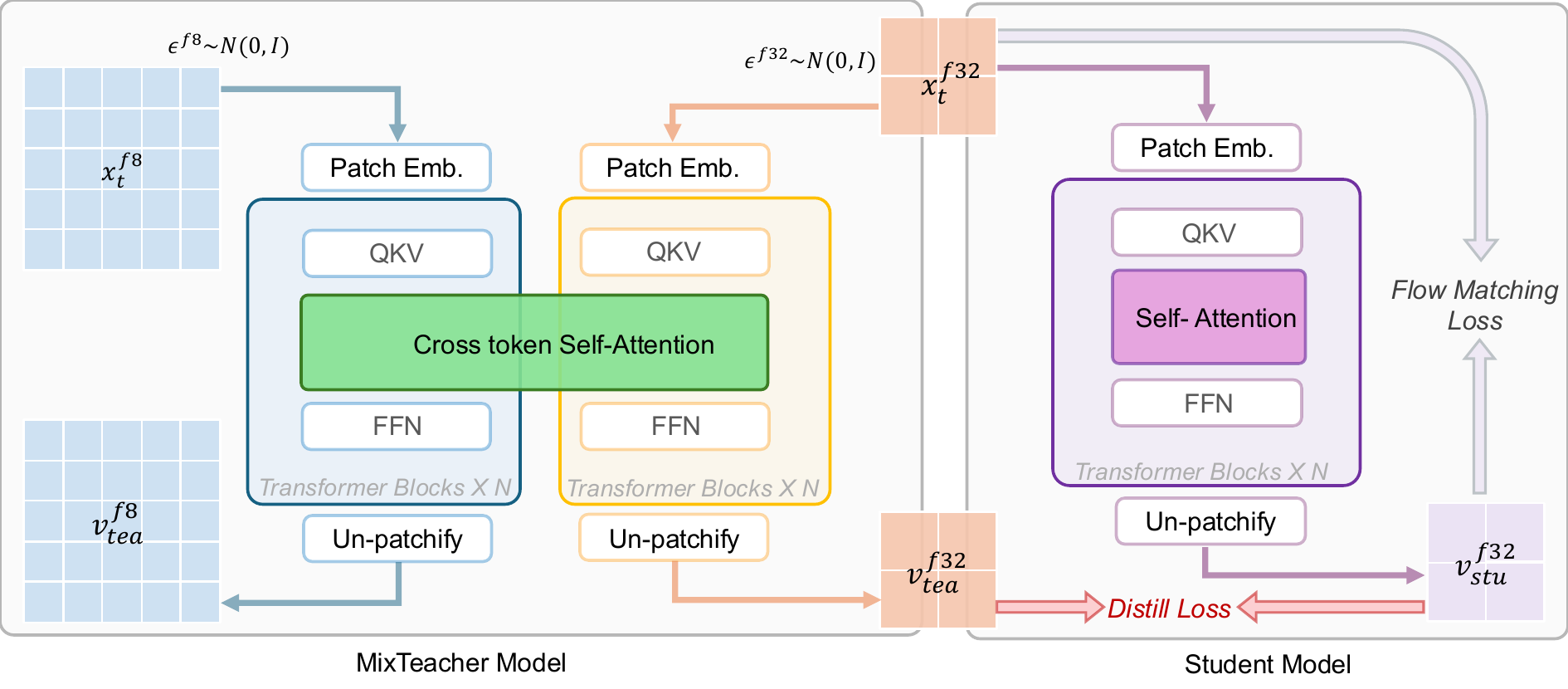}
\caption{\textbf{Overview of UltraDiff.} 
(Left) The \textbf{MixTeacher} models the joint distribution of f8 and f32 latents via a dual-branch DiT with shared attention. It acts as an omniscient supervisor that sees both high-fidelity (f8) and high-efficiency (f32) views of the data. (Right) 
The \textbf{Student} is a standard f32 DiT trained via \textit{Marginal Score Distillation}. It learns to approximate the teacher's optimal denoising direction projected onto the f32 subspace with a flow-matching objective.}
\label{fig:main_method}
\end{figure}

We formulate the problem of high-efficiency, high-fidelity synthesis as joint-latent diffusion learning followed by marginal score distillation. Our goal is to train an efficient student model that operates in the high-compression space  but inherits the generative vector field of the high-fidelity joint distribution.

\textbf{Motivation: Feasibility of Cross-Space Transfer.}
Before detailing our framework, we first empirically validate the hypothesis that the generative representations of f8 (high-fidelity) and f32 (high-efficiency) latent spaces share transferable structural patterns, despite their representation and dimensional mismatch. We conducted a pilot study comparing a baseline f32 DiT model trained from scratch against one initialized with parameters from a pre-trained f8 model (excluding latent-dependent layers). As shown in Fig.~\ref{fig:geneval_iter}, the f8-initialized model consistently outperforms the randomly initialized baseline. This confirms that priors learned in the f8 space are transferable. However, a persistent performance gap remains. This saturation suggests that static parameter initialization is insufficient. This motivates our proposal of MixTeacher, which enables dynamic, interaction-based knowledge transfer via joint latent modeling.

\subsection{Joint-Latent Modeling via MixTeacher}
\label{subsec:mixteacher}

To capture the correlation between high-frequency details (f8) and compact semantics (f32), we introduce MixTeacher, a unified diffusion model parameterized by $\Phi$. We consider the data as tuples in a product space $\mathcal{X}_{\text{joint}} = \mathcal{X}_{\text{f8}} \times \mathcal{X}_{\text{f32}}$. The MixTeacher approximates the joint score function $\nabla \log p(\mathbf{x}^{\text{f8}}, \mathbf{x}^{\text{f32}})$.
To provide reliable cross-space guidance, we design MixTeacher as a mixed-token dual-branch DiT. As illustrated in Fig.~\ref{fig:main_method}, the f8 and f32 branches maintain separate parameters---including QKV projections, FFNs, normalization layers, patch embeddings, and output heads---but share the self-attention mechanism across tokens in each layer. This design enables cross-latent space interaction while respecting the distinct distributions of each space.

Formally, given an input image $\mathbf{I} \in \mathbb{R}^{3 \times H \times W}$, we extract its f8 and f32 latents via respective VAE encoders: $\mathbf{x}^{\mathrm{f8}}_0 \in \mathbb{R}^{C_8 \times \frac{H}{8} \times \frac{W}{8}}$ and $\mathbf{x}^{\mathrm{f32}}_0 \in \mathbb{R}^{C_{32} \times \frac{H}{32} \times \frac{W}{32}}$.
During training, the noisy latents $\mathbf{x}_t^{\mathrm{f8}}$ and $\mathbf{x}_t^{\mathrm{f32}}$ are fed to the MixTeacher.
Inside the model, each branch processes its own tokens using its specific parameters. Crucially, the resulting token sets are concatenated and passed through the shared self-attention at every layer. This enables dense interaction between the two modalities, allowing the f32 tokens to incorporate fine-grained structural details in the f8 tokens.
The model output is formulated as the predicted velocity field $\mathbf{v}$, which is supervised by the target vector field $\mathbf{u}$ derived from the forward flow matching process.

\textbf{Training Objective and Stochastic Routing.}
We train the MixTeacher using a flow-matching objective~\citep{lipman2022flow}. To capture the joint distribution while maintaining generative capabilities of individual branches over the f8 or f32 marginals, we employ a stochastic routing strategy.
Formally, we optimize a weighted objective across three topological configurations defined by a branch indicator mask $(m_{\mathrm{\texttt{f8}}}, m_{\mathrm{\texttt{f32}}}) \in \{(1,1), (1,0), (0,1)\}$, indicating the activation status of the f8 and f32 branches, respectively. The unified loss function is:
\begin{equation}
\mathcal{L} = \mathbb{E}_{t, \mathbf{x},(m_{\mathrm{\texttt{f8}}}, m_{\mathrm{\texttt{f32}}})} \left[\sum_{k \in {\mathrm\{\texttt{f8}, \texttt{f32}\}}} m_k \cdot \|\mathbf{v}_{\Phi}^{k}(\mathbf{x}_t^k) - \mathbf{u}_t^{k} \|^2 \right] ,
\label{eq:unified_loss}
\end{equation}
where $m_k \in \{0, 1\}$ controls whether loss is applied on the specified latent space, $\mathbf{v}_{\Phi}^{k}$ is the vector field predictor parametrized by $\Phi$ and $\mathbf{u}_t^{k}$ is the target vector field.

More specifically, we sample among: (i) \textbf{Joint Mode} ($m_{\texttt{f8}}=1, m_{\texttt{f32}}=1$) while self-attention between two branches is enabled to learn the score jointly; (ii) \textbf{f8-Solo} ($m_{\texttt{f8}}=1, m_{\texttt{f32}}=0$), preserving valid generation in the original f8 space; and (iii) \textbf{f32-Solo} ($m_{\texttt{f8}}=0, m_{\texttt{f32}}=1$), which is crucial for ensuring the generation capability of the f32 branch without conditioning on f8 tokens.
The final optimization objective is the expectation over these routing modes, governed by the sampling probabilities $p_{\text{joint}}, p_{\text{f8}},p_{\text{f32}}$.
This strategy prevents the "co-adaptation" pitfall where the f32 branch might become overly reliant on f8 features, ensuring that the structural knowledge transfer is robust and generalizable.



\subsection{Distilling the Marginal Score}
\label{subsec:distill_strategy}
Once trained, MixTeacher provides a strong teacher signal for supervising the f32 student. Specifically, we aim to distill the knowledge of MixTeacher into a student model $\theta$ that operates \textit{only} in the f32 marginal space, while the student aims to mimic the vector field predicted by the teacher's f32 branch. However, since the student lacks access to the high-fidelity f8 latent $\mathbf{x}_t^{\text{f8}}$ utilized by the teacher, the distillation target is inherently stochastic from the student's perspective. We formulate the objective as minimizing the expected discrepancy over the joint distribution of the f8 and f32 latents:
\begin{equation}
\label{eq:marginal_distill}
\mathcal{L}_{\text{distill}}(\theta)
= \mathbb{E}_{t, (\mathbf{x}^{\text{f32}}_t, \mathbf{x}^{\text{f8}}_t) \sim q_t} \Big[
\big\|
\mathbf{v}_{\theta}(\mathbf{x}^{\text{f32}}_{t}, t) -
\mathbf{v}_{\Phi}^{\text{f32}}(\mathbf{x}^{\text{f32}}_{t}, \mathbf{x}^{\text{f8}}_{t}, t)
\big\|^2
\Big].
\end{equation}

Minimizing this $L_2$ objective inherently forces the student to approximate the conditional expectation of the teacher's output given the $f32$ observation, i.e., 
the optimal estimator $\mathbf{v}_{\theta}^*$ is thus given by:
\begin{equation}
\label{eq:conditional_expectation}
\mathbf{v}_{\theta}^*(\mathbf{x}_t^{\text{f32}}, t) = \mathbb{E}_{q(\mathbf{x}^{\text{f8}} | \mathbf{x}^{\text{f32}})} \Big[ \mathbf{v}_{\Phi}^{\text{f32}}(\mathbf{x}_t^{\text{f32}}, \mathbf{x}_t^{\text{f8}}, t) \Big].
\end{equation}

The student learns the marginal vector field projected from the joint space. By aggregating supervision over the coherent f8 priors provided by the teacher, this projection acts as a structural filter that effectively ``integrates out'' the ambiguity and stochastic aliasing artifacts inherent to the f32 space. The student is thus guided toward a structurally consistent manifold rather than fitting the noisy, underspecified f32 distribution directly.

Crucially, this projection does not result in a blurry average. As visualized in Fig.~\ref{fig:mixteacher_vis}, in the mixed routing mode, the output of the teacher's f32 branch is visually sharp and indistinguishable from the f8 branch. This indicates that the teacher's attention sharing mechanism successfully resolves ambiguity, collapsing the conditional distribution into a highly peaked, determinate signal. Consequently, the distillation target $\mathbf{v}_{\Phi}^{\text{f32}}$ serves as a structurally determinate vector field, providing a clean supervision signal that enables the student to recover fine details despite the high compression ratio. We train the student model using a composite objective of distillation loss and flow matching loss: \begin{equation} \label{eq:student_total} \mathcal{L}_{\text{total}} = \lambda \mathcal{L}_{\text{distill}} + \mathbb{E}_{t, \mathbf{x}} \left[\| \mathbf{v}_{\theta}(\mathbf{x}^{\text{f32}}_{t}) - \mathbf{u}_t^{\text{f32}} \|^2 \right]. 
\end{equation}


\begin{wraptable}{r}{0.48\linewidth}
\caption{\textbf{Controlled comparison under the same framework.}
All f32 variants use the same f32 VAE and DiT backbone. Initialization alone transfers part of the f8 prior, whereas marginal score distillation recovers most of the left gap.}
\centering
\setlength{\tabcolsep}{4pt}
\renewcommand{\arraystretch}{1.30}
\resizebox{\linewidth}{!}{
\begin{tabular}{lcccc}
\toprule
\rowcolor{gray!10}
\textbf{Method} & \textbf{CLIP$\uparrow$} & \textbf{IR$\uparrow$} & \textbf{GenEval$\uparrow$} & \textbf{Token} \\
\midrule
f32 from scratch   & 27.21 & 0.70 & 0.54 & $\frac{H}{32}\times \frac{W}{32}$ \\
f32 after f8 initialization  & 28.91 & 0.91 & 0.61 & $\frac{H}{32}\times \frac{W}{32}$ \\
UltraDiff f32 student & 29.27 & 0.98 & 0.72 & $\frac{H}{32}\times \frac{W}{32}$ \\
f8 anchor       & 29.25 & 1.08 & 0.73 & $\frac{H}{8}\times \frac{W}{8}$ \\
\bottomrule
\end{tabular}
}
\label{tab:f32train}
\vspace{-10pt}
\end{wraptable}

This comparison separates the effect of the proposed training procedure from the strength of the pretrained f8 anchor. Directly reusing the f8 parameters improves the f32 model from 0.54 to 0.61 GenEval, but leaves a substantial gap to the 0.73 f8 anchor. With the same f8 initialization, f32 VAE, and DiT backbone, UltraDiff reaches 0.72. The gain from 0.61 to 0.72 therefore cannot be explained by initialization or VAE quality alone and instead measures the contribution of joint-latent modeling and marginal score distillation.

\section{Experiment}

\subsection{Implementation Details}
We employ a Diffusion Transformer (DiT) architecture that models text and image tokens as a unified sequence. Specifically, text embeddings (via T5-XXL~\citep{raffel2020exploring}) and image latents are concatenated along the sequence dimension and processed through a shared bidirectional transformer backbone.
We use a patch size of $1$ for image embeddings.
For latent encoding, we utilize high-capacity VAEs with latent channel dimensions of $C_{\text{f8}}=16$ and $C_{\text{f32}}=64$, corresponding to $8\times$ and $32\times$ downsampling factors, respectively.
We apply channel-wise unit-variance normalization to both f8 and f32 latents before they enter the DiT backbone. 
We first train a DiT in the f32 latent space under two settings: (i) from scratch, and (ii) initialized from the f8 model's weights, except for the shape-dependent patch embedding and output heads, which are randomly initialized.
For MixTeacher, the f8 branch is initialized from a strong pretrained f8 T2I model, and the f32 branch is initialized from the previously trained f32 model with f8 initialization. 
We train the teacher at 512 and 1024 resolutions with varying aspect ratios.
All UltraDiff stages use 64 NVIDIA A100 GPUs with FSDP. After filtering and deduplication, the available corpus contains approximately 0.37B text--image pairs. Table~\ref{tab:training_resources} reports the optimization steps and aggregate compute for each stage. The three stages consume 463.0 GPU-days in total, about 26\% of the 1792 GPU-days required to train our lower-quality f32 baseline directly. These costs treat the fixed f8/f32 VAEs and pretrained f8 anchor as available components; further optimization settings are provided in the appendix.
\vspace{-5pt}

\begin{table*}[t]
\centering
\caption{\textbf{Stage-wise training resources.} Compute is aggregated over 64 NVIDIA A100 GPUs.}
\label{tab:training_resources}
\small
\setlength{\tabcolsep}{4pt}
\renewcommand{\arraystretch}{1.08}
\begin{tabular}{lccc}
\toprule
\textbf{Stage} & \textbf{Steps} & \textbf{Compute} & \textbf{Output} \\
\midrule
f32 transfer & 120K & 66.7 GPU-days & f32 initialization \\
MixTeacher training & 5K & 29.6 GPU-days & Joint f8--f32 teacher \\
Student distillation and scaling & 50K + 10K & 366.7 GPU-days & Standalone f32 student \\
\midrule
\rowcolor{gray!10}
UltraDiff adaptation total & 185K & 463.0 GPU-days & Final UltraDiff model \\
\bottomrule
\end{tabular}
\end{table*}

\subsection{Ablation Study}

\label{sec:exp_abla}
\textbf{Evaluation criteria. } We evaluate models using GenEval~\citep{ghosh2023geneval}, CLIP similarity~\citep{radford2021learning}, and ImageReward (IR)~\citep{xu2023imagereward} at a resolution of $512\times 512$ for fast iteration. CLIP and IR are computed on a 3k subset of MJHQ-30k~\citep{li2024playground} to further reduce evaluation cost.

\noindent\textbf{Train in f32 space.}
We train the T2I model in the f32 latent space for 120k iterations until convergence.
The results show that the model with f8 initialization exhibits much better performance than the model trained from scratch. As shown in Figure~\ref{fig:geneval_iter}, the model initialized from f8 (green curve) consistently outperforms the scratch-trained model (blue curve) according to GenEval~\citep{ghosh2023geneval} throughout training. Table~\ref{tab:f32train} reports similar gains on CLIP and IR metrics, showing the benefit of transferring parameters from the f8 model.
However, simple initialization still leaves a large gap compared with the f8 model, as shown in Figure~\ref{fig:geneval_iter} and Table~\ref{tab:f32train}. This suggests that, in the highly compressed f32 space, good initialization helps but is not sufficient. Even with f8 initialization, optimization remains difficult, and the model cannot fully match the f8 teacher under a purely data-driven objective.
Leveraging the strong f8 model, we distill its behavior to enhance f32 performance.
\begin{figure}[t]
\centering
\begin{minipage}[t]{0.54\textwidth}
  \vspace{7pt}
  \centering
  \includegraphics[width=\linewidth]{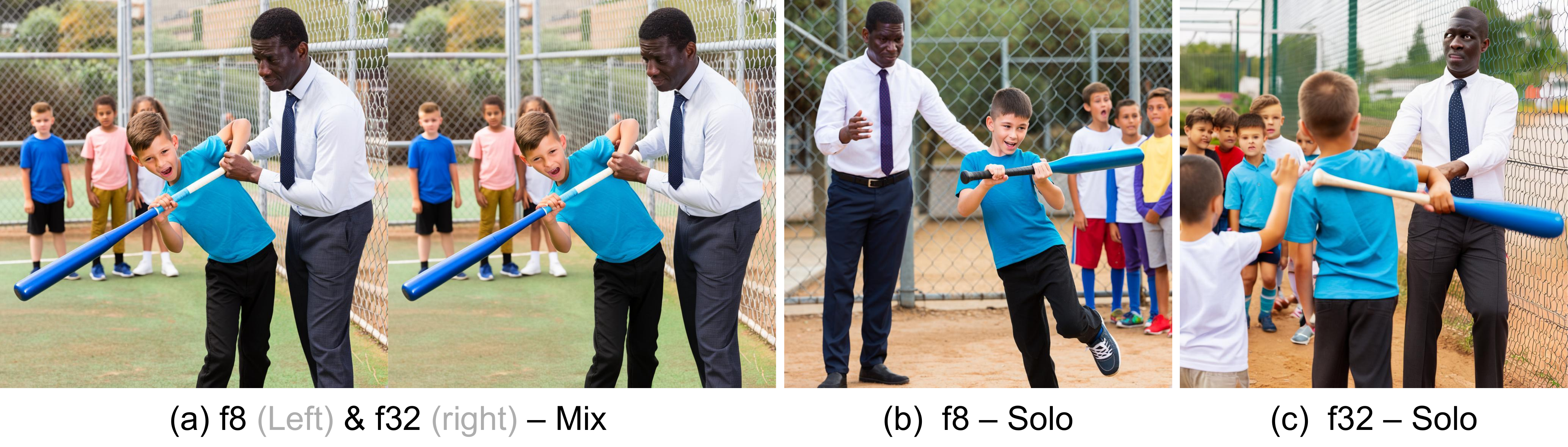}
  \vspace{-15pt}
  \caption{Visual results of different inference modes of the MixTeacher.}
  \vspace{-10pt}
  \label{fig:mixteacher_vis}
\end{minipage}
\hfill
\begin{minipage}[t]{0.43\textwidth}
  \vspace{0pt}
  \centering
  \captionof{table}{MJHQ-3K and GenEval results for different MixTeacher modes. The f32 branch in mixed mode matches the performance of the base f8 model.}
  \setlength{\tabcolsep}{7pt}
  \renewcommand{\arraystretch}{1.05}
  \resizebox{\linewidth}{!}{
  \begin{tabular}{lcccc}
  \toprule
  \rowcolor{gray!10}
  \textbf{Method} & \textbf{CLIP$\uparrow$} & \textbf{IR$\uparrow$} & \textbf{GenEval$\uparrow$} & \textbf{Token} \\
  \midrule
  f8 \& f32 Mixed (f8 Branch)  & 29.27 & 1.05 & 0.74 & $\frac{H}{8}\times \frac{W}{8}$  \\
  f8 \& f32 Mixed (f32 Branch) & 29.24 & 1.02 & 0.74 & $\frac{H}{32}\times \frac{W}{32}$ \\
  \midrule
  f8-Solo  & 29.26 & 1.06 & 0.73 & $\frac{H}{8}\times \frac{W}{8}$ \\
  \midrule
  f32-Solo & 28.88 & 0.85 & 0.62 & $\frac{H}{32}\times \frac{W}{32}$ \\
  \bottomrule
  \end{tabular}
  }
  \label{tab:teacher_infer}
\end{minipage}
\end{figure}


\begin{figure}[t]
\centering

\begin{minipage}[t]{0.56\linewidth}
    \vspace{0pt}
    \centering
    \includegraphics[width=\linewidth]{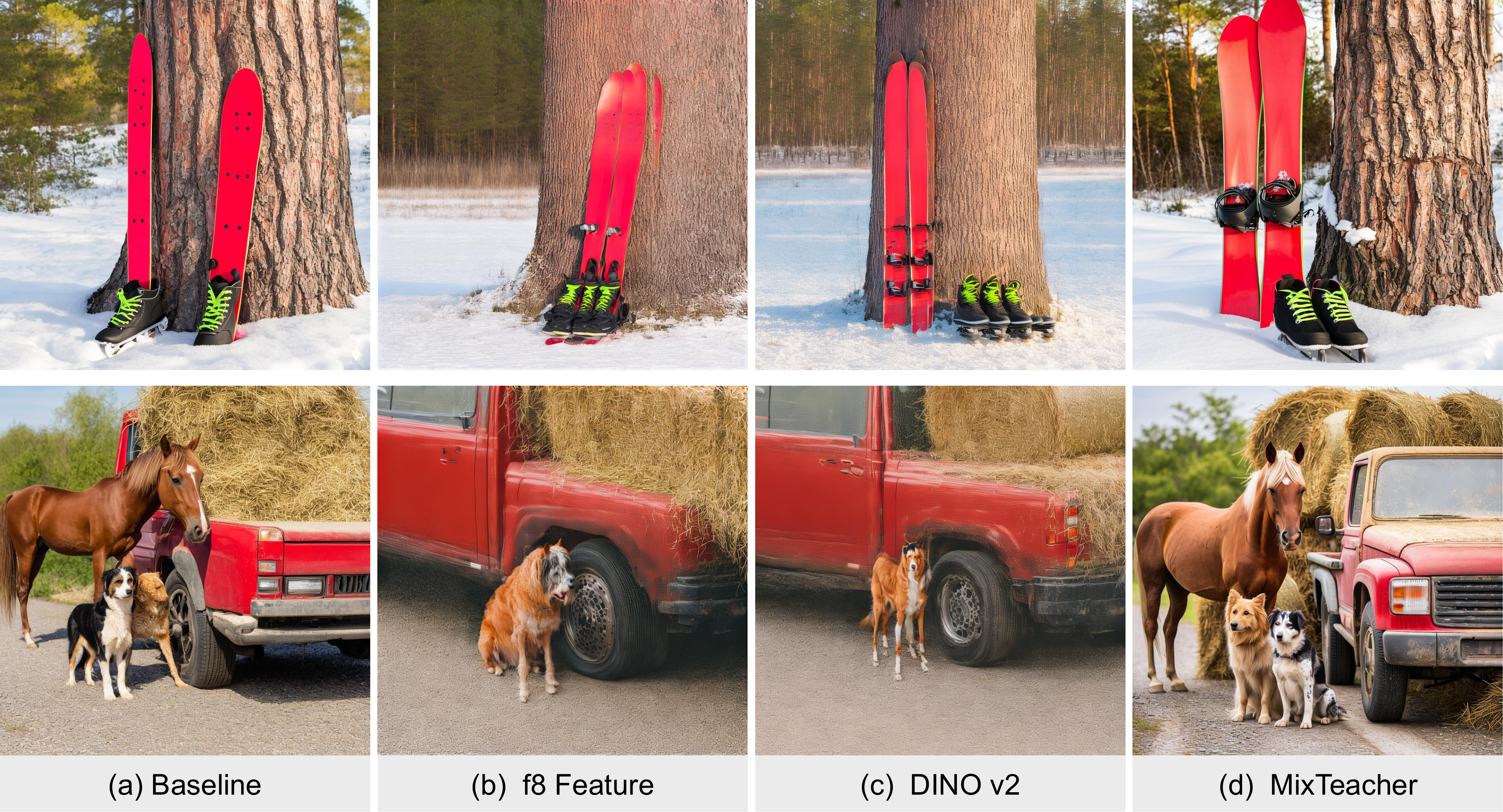}
    \caption{Visual comparison of (a) the f8 base model, 
    (b) f32 model distilled from an f8 teacher~\citep{ren2025turbo2k}, 
    (c) f32 model distilled from DINOv2 features via REPA~\citep{yu2024representation}, 
    and (d) f32 model distilled from our MixTeacher.}
    \label{fig:exp_distill_vis}
\end{minipage}
\hfill
\begin{minipage}[t]{0.43\linewidth}
    \vspace{5pt}
    \centering
    \begin{minipage}[t]{\linewidth}
        \vspace{0pt}
        \captionof{table}{Results on MJHQ-3K and GenEval for different distillation methods.}
        \centering
        \setlength{\tabcolsep}{5pt}
        \renewcommand{\arraystretch}{1.0}
        \resizebox{\linewidth}{!}{
        \begin{tabular}{lccc}
        \toprule
        \rowcolor{gray!10}
        \textbf{Teacher} & \textbf{CLIP$\uparrow$} & \textbf{IR$\uparrow$} & \textbf{GenEval$\uparrow$} \\
        \midrule
        Baseline & 28.91 & 0.91 & 0.61 \\
        DINO v2~\citep{leng2025repa} & 27.53 & 0.06 & 0.33 \\
        f8 model~\citep{ren2025turbo2k} & 27.87 & 0.05 & 0.31 \\
        MixTeacher (Ours) & 29.27 & 0.98 & 0.72 \\
        \bottomrule
        \end{tabular}
        }
        \label{tab:distill_strategy}
    \end{minipage}

    \vspace{8pt}

    \begin{minipage}[t]{\linewidth}
        \vspace{0pt}
         \captionof{table}{Impact of routing strategy on GenEval across inference modes.}
        \centering
        \renewcommand{\arraystretch}{0.9}
        \setlength{\tabcolsep}{3.5pt}
        \scriptsize
        \resizebox{\linewidth}{!}{
        \begin{tabular}{ccc|cccc}
        \toprule
        \multicolumn{3}{c}{\textbf{Routing Strategy}} & \multicolumn{4}{c}{\textbf{Inference Mode (GenEval $\uparrow$)}} \\
        \cmidrule(lr){1-3} \cmidrule(lr){4-7}
        \boldmath$p_{\text{mix}}$ & \boldmath$p_{\text{f32}}$ & \boldmath$p_{\text{f8}}$
        & \textbf{f8-Solo} & \textbf{f32-Solo} & \textbf{Mix(f8)} & \textbf{Mix(f32)} \\
        \midrule
        1.0 & 0.0 & 0.0 & 0.56 & 0.28 & 0.68 & 0.68 \\
        0.5 & 0.0 & 0.5 & 0.65 & 0.15 & 0.65 & 0.64 \\
        \midrule
        \rowcolor{gray!10}
        0.4 & 0.4 & 0.2 & 0.73 & 0.62 & 0.74 & 0.74 \\
        \bottomrule
        \end{tabular}
        }
        \label{tab:routing_ablation}
    \end{minipage}

\end{minipage}
\vspace{-10pt}
\end{figure}

\noindent\textbf{Exploration of MixTeacher Routing Strategy.}
To identify the optimal strategy for training the MixTeacher, we evaluate three routing configurations in Table~\ref{tab:routing_ablation}. First, exclusive joint training ($p_{joint} = 1.0$) performs poorly; the lack of single-branch constraints causes the f32 branch to exploit f8 shortcuts, leading to representation collapse. Second, mixing only Joint and f8-Solo modes (e.g., $p_{joint}=0.5, p_{f8}=0.5$) also fails, as the f32 branch remains under-trained, creating a bottleneck that drags down the joint performance. These results underscore a critical insight: maintaining the independent generative capability of each branch is a prerequisite for learning a robust joint distribution. Consequently, we adopt a balanced configuration ($p_{joint}=0.4, p_{f32}=0.4, p_{f8}=0.2$). This strategy explicitly enforces the marginal capability of the f32 branch, ensuring it provides a high-quality supervision target. We assign a lower weight to f8-Solo ($0.2$) since the strong pre-trained f8 branch requires less tuning, allowing us to allocate more budget to the f32 and joint objectives.

\noindent\textbf{Analysis of MixTeacher Inference.}
Having determined the suitable routing configuration, we further analyze the inference behavior of the MixTeacher (Table~\ref{tab:teacher_infer}).
By comparing the inference modes, we observe that the Mixed Mode effectively bridges the gap between the coarse f32 latent space and the high-fidelity target. While the f32-Solo baseline ($0.62$ GenEval) yields suboptimal performance due to the aggressive compression, the Mixed Mode ($0.74$ GenEval) demonstrates that the teacher successfully utilizes the f8 context to resolve structural ambiguities. Crucially, as visualized in Figure~\ref{fig:mixteacher_vis}, the images generated by the f32 and f8 branches in Mixed Mode are visually indistinguishable, matching in both high-frequency details and semantic alignment. This indicates that although the two branches reside in distinct latent spaces, the MixTeacher successfully projects the high-fidelity priors into the compressed f32 representation. Consequently, this f32 output serves as a high-quality, structurally determinate distillation signal directly within the student's native space.

\textbf{Distillation strategy}
We explore four strategies in Table \ref{tab:distill_strategy} and Fig. \ref{fig:exp_distill_vis}:
(1) Baseline --- an f32 model initialized from the f8 model (no distillation).
(2) DINOv2 teacher --- align the DiT's internal features to DINOv2 following REPA~\citep{leng2025repa}.
(3) f8 feature teacher --- align f32 features to those from the f8 model following~\citet{ren2025turbo2k}.
(4) MixTeacher (ours) --- process f8 and f32 tokens together and use the f32 branch as the distillation signal.
External semantic guidance from DINOv2 brings little gain. Our baseline has already been pre-trained in f8 and then fine-tuned in f32, so semantic priors are strong; additional supervision from a generic image encoder is redundant and not matched to the diffusion objective. Moreover, feature alignment to DINOv2 requires grid upsampling from f32, which induces checkerboard/ringing even with learned convolutional mapping. The problem is amplified under variable aspect ratios (tall/wide samples), where DINOv2 offers limited spatial guidance for the diffusion trajectories.
Directly aligning to the f8 feature grid performs similarly or worse for the same reason: the f8 lattice is $4\times$ finer per dimension than f32, so supervision again depends on learned upsampling across mismatched token layouts, introducing aliasing and bias.
In contrast, ours supervises in the f32 space using the f32 branch's v-prediction. This avoids cross-grid interpolation, matches the denoising objective, and yields sharper details and better text--image consistency, as seen qualitatively in Figure~\ref{fig:exp_distill_vis} and quantitatively in Table~\ref{tab:distill_strategy}.

\noindent\textbf{Distillation versus high-resolution training.} To isolate the effect of marginal score distillation, we train a same-backbone control from the same f8-initialized f32 checkpoint, using identical high-resolution scaling but without distillation. During scaling, 1K samples use both distillation and flow matching, while 2K–4K samples use flow matching only. As shown in Table~\ref{tab:distill_highres_control}, UltraDiff improves DPG, GenEval, and HPSv3 by 11.38, 0.078, and 1.71 at 1K, respectively, while also achieving better FID, aesthetic score, and CLIP similarity at 4K. These results highlight the complementary roles of distillation in transferring the f8 prior and high-resolution flow matching in adapting to larger spatial scales.

\begin{table*}[t]
\centering
\caption{\textbf{Controlled attribution of the high-resolution gains.} Both models use the same f8-initialized f32 backbone and the same mixed-resolution scaling stage. DPG, GenEval, and HPSv3 are evaluated at 1K; FID, aesthetic score, and CLIP similarity are evaluated at 4K.}
\label{tab:distill_highres_control}
\small
\setlength{\tabcolsep}{5pt}
\renewcommand{\arraystretch}{1.1}
\resizebox{0.7\linewidth}{!}{%
\begin{tabular}{lcccccc}
\toprule
\textbf{Model} & \textbf{DPG}$\uparrow$ & \textbf{GenEval}$\uparrow$ & \textbf{HPSv3}$\uparrow$ & \textbf{FID}$\downarrow$ & \textbf{Aesthetic}$\uparrow$ & \textbf{CLIP}$\uparrow$ \\
\midrule
No distillation + high-res & 74.02 & 0.602 & 7.68 & 149.59 & 6.16 & 29.31 \\
\rowcolor{gray!10} UltraDiff & \textbf{85.40} & \textbf{0.680} & \textbf{9.39} & \textbf{146.60} & \textbf{6.34} & \textbf{30.14} \\
\bottomrule
\end{tabular}}
\end{table*}

\begin{figure}[t]
  \centering
   \includegraphics[width=0.9\linewidth]{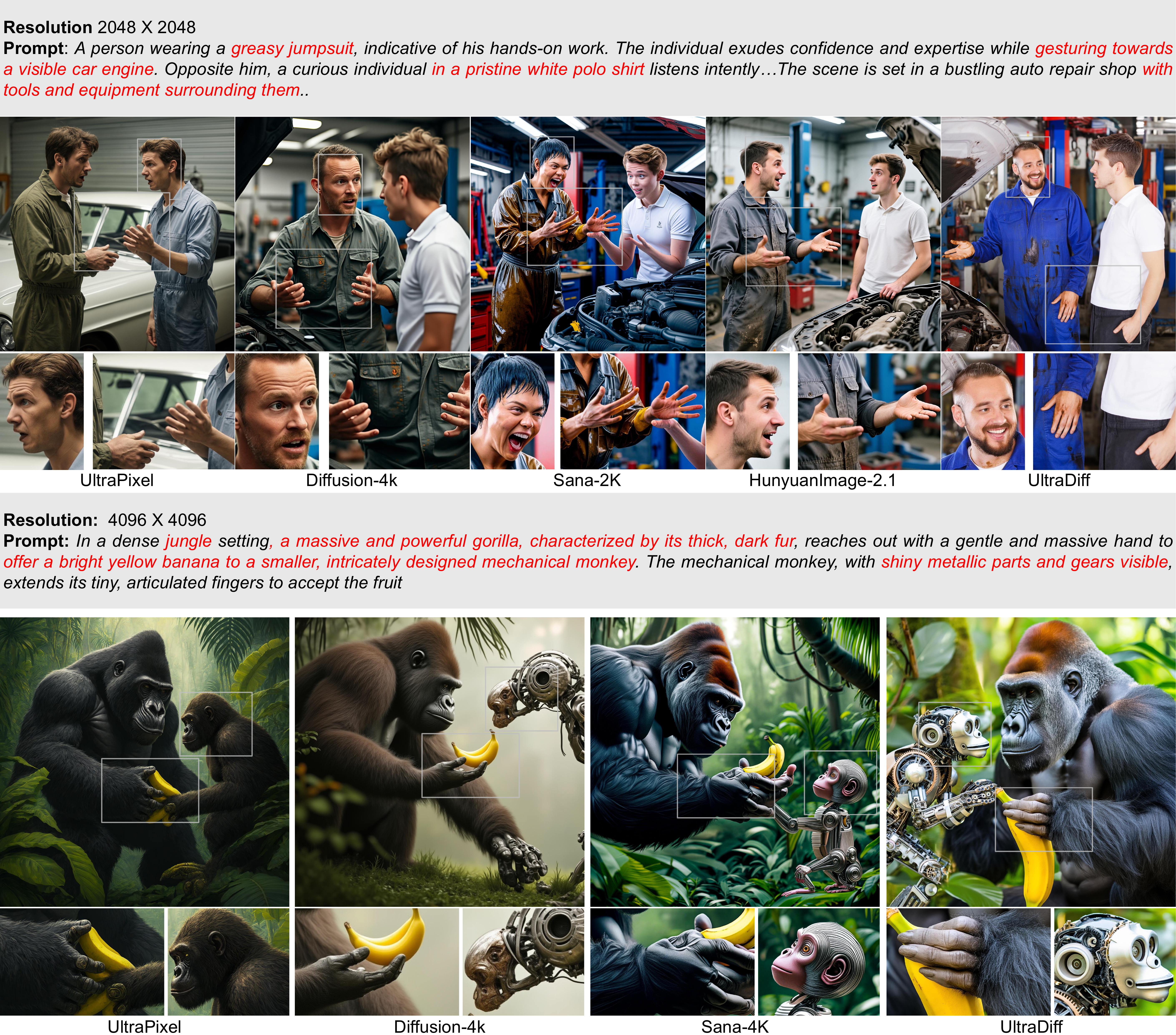}
   \caption{High-resolution image synthesis comparison at 2K and 4K. UltraDiff produces sharper details, cleaner textures, and stronger text--image alignment than competing high-resolution models.}
   \label{fig:compare_2k4k}
  \vspace{-20pt}
\end{figure}


\label{sec:experiment}

\begin{table}[t]
\centering
\caption{Comparison with other T2I models on higher resolution of $2048\times 2048$ and $4096\times 4096$.}
\setlength{\tabcolsep}{7pt}
\renewcommand{\arraystretch}{1.2}
\resizebox{1\linewidth}{!}{
\begin{tabular}{llcccccccc}
\toprule
\rowcolor{gray!10}\textbf{Resolution} & \textbf{Method} & \textbf{Params(B)} & \textbf{Latent(\text{$C\times H \times W$)}}&\textbf{Latency(s) / Step} & \textbf{FID}$\downarrow$ & \textbf{pFID}$\downarrow$ & \textbf{Aesthetic}$\uparrow$& \textbf{CLIP}$\uparrow$\\
\midrule
\multirow{5}{*}{$2048^2$} & Ultrapixel~\citep{ren2024ultrapixel} &5.1&$16\times\frac{H}{32}\times \frac{W}{32}$  & 12 / 20 & 46.60 & 43.03 &6.24&30.21 \\
                         & Diffusion-4k~\citep{zhang2025diffusion4k} & 12&$16\times\frac{H}{42}\times \frac{W}{42}$ &23 / 50 & \textbf{37.79} & \emph{42.43}  &6.28&29.21 \\
                         & Sana-2K ~\citep{xie2024sana}        &1.6&$32\times\frac{H}{32}\times \frac{W}{32}$ & \textbf{5.8 / 20} &45.99  & 42.99 &6.18 &30.13\\
                         & Hunyuan-2.1~\citep{hunyuanimage21_github}  &17&$64\times\frac{H}{32}\times \frac{W}{32}$ & 82 / 50 & 43.45 & 48.72 &\textbf{6.44}&\emph{30.17} \\
\rowcolor{gray!10}       & Ours         & 2.0&$64\times\frac{H}{32}\times \frac{W}{32}$ & \emph{6.2 / 20} &\emph{39.52}  & \textbf{42.17} &\emph{6.40}&\textbf{30.53} \\
\midrule
\multirow{4}{*}{$4096^2$} & Ultrapixel~\citep{ren2024ultrapixel}   & 5.1&$16\times\frac{H}{42}\times \frac{W}{42}$ & 73 / 20  & 155.17 & 57.14 &5.91&27.39 \\
                         & Diffusion-4k~\citep{zhang2025diffusion4k} & 12.0&$16\times\frac{H}{32}\times \frac{W}{32}$ & 122 / 50 &  162.94&75.70  &5.68 & 27.20\\
                         & Sana-4K  ~\citep{xie2024sana}    &  1.6&$32\times\frac{H}{32}\times \frac{W}{32}$ & \textbf{21.3 / 20}   &\emph{149.20}  & \emph{56.25} & \emph{6.12}&\emph{29.97}\\
\rowcolor{gray!10}       & Ours         & 2.0&$64\times\frac{H}{32}\times \frac{W}{32}$ & \emph{28.3 / 20} & \textbf{146.60} & \textbf{54.27} & \textbf{6.34}&\textbf{30.14}\\
\bottomrule
\end{tabular}}
\label{tab:compare_2k4k}
\vspace{-15pt}
\end{table}

\begin{table}[t]
\caption{Comparison with other T2I models on $1024^2$.}
\centering
\setlength{\tabcolsep}{7pt}
\renewcommand{\arraystretch}{1.2}
\resizebox{1\linewidth}{!}{
\begin{tabular}{lccccc}
\toprule
\rowcolor{gray!10}\textbf{Method} & \textbf{Token} &\textbf{Params(B)} &  \textbf{DPG}$\uparrow$ & {\textbf{GenEval}$\uparrow$} & {\textbf{HPSv3}$\uparrow$}\\
\midrule
SDXL~\citep{podell2023sdxl} & $\frac{H}{8}\times\frac{W}{8}$ &2.6  &  74.6 & 0.55 &7.87\\
PlayGroundv2.5~\citep{li2024playground} &$\frac{H}{8}\times \frac{W}{8}$& 2.6 & 75.5 & 0.56 & \emph{9.90}\\
\hline
PixArt$\Sigma$~\citep{chen2024pixart}&$\frac{H}{16}\times\frac{W}{16}$ & 0.6  & 80.5 & 0.54 &9.20\\
SD3-medium~\citep{esser2024scaling} & $\frac{H}{16}\times \frac{W}{16}$ & 2.0 & 84.1 &0.62  &9.08\\
FLUX-dev~\citep{FLUX} &$\frac{H}{16}\times \frac{W}{16}$& 12 & 83.8 & 0.67 & \textbf{10.15} \\
FLUX-schnell~\citep{FLUX} &$\frac{H}{16}\times \frac{W}{16}$&12   & 84.8 & \textbf{0.71} & 9.37\\
\hline
Sana~\citep{xie2024sana}  &$\frac{H}{32}\times \frac{W}{32}$& 1.6  & \emph{84.5} & 0.66 & 8.92\\
\rowcolor{gray!10}UltraDiff (Ours)&$\frac{H}{32}\times \frac{W}{32}$&2.0& \textbf{85.4} & \emph{0.68}  & 9.39\\
\bottomrule
\end{tabular}
}
\label{tab:compare_1024}
\vspace{-20pt}
\end{table}

\subsection{Comparative Study}
\vspace{-5pt}
\textbf{Evaluation Criteria} We evaluate our UltraDiff and other methods using DPG Bench~\citep{hu2024ella}, GenEval~\citep{ghosh2023geneval} and human preference score (HPSV3)~\citep{ma2025hpsv3widespectrumhumanpreference}.
For quantitative evaluation, we adopt the protocols from Diffusion-4K~\citep{zhang2025diffusion4k}. FID is computed on the entire Aesthetic-4K~\citep{zhang2025diffusion4k} evaluation set, with images resized to $299 \times 299$. To assess high-frequency details, we report pFID, calculated by extracting 4 random non-overlapping patches from each ground truth and generated image. 
\noindent\textbf{Results at $1024^2$.} We compare UltraDiff with complete T2I systems based on f8 VAEs~\citep{FLUX, podell2023sdxl, chen2024pixart, li2024playground} and with the f32-based Sana system~\citep{xie2024sana}. Table~\ref{tab:compare_1024} shows that UltraDiff offers a strong practical quality--efficiency trade-off, ranking at or near the top among models of similar size on DPG and GenEval while remaining competitive on HPSv3.
These cross-system results provide deployment-oriented context rather than a controlled causal comparison: the systems differ in backbone architecture, parameter count, training data, text encoders, autoencoders, optimization recipes, and inference infrastructure. We therefore establish the contribution of UltraDiff using the within-framework controls in Tables~\ref{tab:f32train} and~\ref{tab:distill_highres_control}, where the relevant f32 models share the same VAE, DiT backbone, initialization, and high-resolution training protocol.

\noindent\textbf{Results of $2048^2$ and $4096^2$.}
UltraDiff uses a high VAE compression ratio ($\frac{H}{32}\times\frac{W}{32}$), which enables 2K and 4K synthesis within one checkpoint that supports multiple resolutions natively. 
We compare 2K and 4K results against methods capable of high-resolution output, including approaches that scale pretrained models through architectural or inference-time adjustments, such as UltraPixel and Diffusion-4K, and systems that natively support such resolutions, such as Sana-2K, Sana-4K, and HunyuanImage-2.1, as shown in Figure~\ref{fig:compare_2k4k} and Table~\ref{tab:compare_2k4k}. 
At 2K the post adjusted methods are competitive and UltraDiff maintains strong metrics with fast inference, while at 4K these adjusted methods degrade in quality, for example Ultrapixel struggles to preserve semantic alignment and Diffusion-4K begins to produce blurred results. Hunyuan-2.1, which employs a larger parameter budget with the same f32 compression, also performs well, suggesting that high compression scales effectively when paired with sufficient capacity. 
Both Sana and UltraDiff are efficient at high resolution. Sana is faster because it employs linear attention, whereas UltraDiff uses self-attention at a similar parameter scale and compression ratio. Under the reported system-level evaluation, UltraDiff obtains stronger overall image-quality and semantic-alignment metrics. Because the two systems differ in more than the attention mechanism, we do not interpret this comparison as isolating the effect of any single component.

\begin{wraptable}{r}{0.48\linewidth}
\caption{\textbf{Controlled inference efficiency.} Time per image with 20 sampling steps on one NVIDIA A100 GPU.}
\centering
\setlength{\tabcolsep}{6pt}
\renewcommand{\arraystretch}{1.1}
\resizebox{\linewidth}{!}{%
\begin{tabular}{lccc}
\toprule
\textbf{Resolution} & \textbf{f8 anchor} & \textbf{UltraDiff} & \textbf{Speedup} \\
\midrule
$1024^2$ & 27.3s & 3.2s & $8.5\times$ \\
$2048^2$ & 249.8s & 6.2s & $40.3\times$ \\
$4096^2$ & OOM & 28.3s & -- \\
\bottomrule
\end{tabular}
}
\label{tab:inference_efficiency}
\vspace{-8pt}
\end{wraptable}

\noindent\textbf{Controlled inference efficiency.}
We compare the final f32 student with the f8 anchor using 20 sampling steps on a single NVIDIA A100 GPU. As shown in Table~\ref{tab:inference_efficiency}, f32 compression reduces the time per image from 27.3s to 3.2s at $1024^2$ ($8.5\times$) and from 249.8s to 6.2s at $2048^2$ ($40.3\times$). At $4096^2$, the f8 model runs out of memory, whereas UltraDiff completes generation in 28.3s. Because both models use the same sampling setup, this comparison isolates the efficiency gained from the smaller f32 latent grid.


\label{sec:exp_compare}


\vspace{-5pt}
\section{Conclusion}
\vspace{-5pt}
In this work, we presented UltraDiff, a novel distillation framework designed to enable high-fidelity image synthesis within highly compressed latent spaces (e.g., f32).
To overcome the inherent difficulty of training in such compact representations, we introduced MixTeacher, a dual-branch architecture that establishes a direct, grid-aligned supervision signal for the student model.
By distilling the marginal vector field from a high-quality anchor directly into the target latent space, our method effectively transfers complex structural details.
Extensive experiments demonstrate that UltraDiff substantially narrows the quality gap between f8 and f32 generation in the evaluated DiT pipeline while retaining the inference efficiency of f32 architectures. With 20 sampling steps on a single A100 GPU, UltraDiff generates a 2K image in 6.2 seconds, a $40.3\times$ speedup over the f8 anchor, and remains practical at 4K where the anchor runs out of memory.
We hope this streamlined distillation paradigm paves the way for more scalable and efficient generative models in the future.


\textbf{Acknowledgment.} This work is supported by the National Natural Science Foundation of China (Project No.82572383), Adobe Research Gift Grant, and Guangdong Basic and Applied Basic Research Foundation (No.2025A1515140021).

\textbf{Impact Statement.} UltraDiff enables efficient, high-quality text-to-image generation in highly compressed latent spaces, potentially reducing the computational and energy costs of 2K–4K image synthesis and broadening access to high-resolution generative models. However, this efficiency may also facilitate the creation of misleading or harmful content. We therefore encourage responsible deployment with appropriate safeguards, including safety evaluation, watermarking, and provenance tracking.

\bibliography{main}
\bibliographystyle{plainnat}

\newpage
\appendix
\onecolumn

\begin{center}
{\LARGE \textbf{Supplementary Material}}\\[0.5em]
{\large \textbf{UltraDiff: Transferring High-Fidelity Priors to Compressed Latent Spaces for High-resolution Image Generation}}
\end{center}

This supplementary material provides comprehensive implementation details and additional results to support the main paper. We organize the content as follows:
\begin{itemize}
    \item \textbf{Section~\ref{app:limitations}} discusses limitation and future work.
    \item \textbf{Section~\ref{suppsec:implement}} provides implementation and reproducibility details, including training protocols, data/model disclosure, and the open-source plan.
    \item \textbf{Section~\ref{app:model_details}} provides the specific architectural specifications and hyperparameters for the DiT backbone, VAEs, and training settings.
    \item \textbf{Additional Visual Results.} Due to the strict file size limit of the PDF submission, we provide extensive visual comparisons and a gallery of high-resolution results in the \textbf{separate attached supplementary files}.
\end{itemize} 

\section{Limitations and Future Work}
\label{app:limitations}
While UltraDiff demonstrates the effectiveness of joint-latent distillation for efficient high-fidelity synthesis, several limitations define the scope of the current evidence.

\textbf{Architecture scope.}
We evaluate UltraDiff on one autoencoder--DiT pipeline. The method only adds latent-specific branches and shared attention during teacher training, and the distilled student remains a standard single-branch DiT. This makes the framework compatible in principle with other DiT-based pipelines, but this architectural compatibility should not be interpreted as empirical validation across backbones. Substantially different denoising architectures remain untested.

\textbf{Dependence on pretrained components and multi-stage training.}
UltraDiff assumes access to a strong f8 text-to-image anchor and fixed f8/f32 VAEs. Its f32-transfer, MixTeacher, and distillation stages add training complexity, although their combined adaptation cost is lower than directly training our f32 baseline (Section~\ref{suppsec:implement}). This complexity is confined to training; inference uses only the final single-branch f32 student.

\textbf{Independent latent spaces.}
The f8 and f32 VAEs are trained independently and are frozen during UltraDiff training. MixTeacher therefore learns their relationship at the diffusion vector-field level rather than enforcing explicit compatibility between their latent spaces. Combining UltraDiff with compatible-VAE training~\citep{yi2025finestructure,zhao2024cvvae} may further improve transfer.

\textbf{Reproducibility and data access.}
The training data and f8 anchor are proprietary, which limits exact reproduction of the reported checkpoint. We provide the architecture, objectives, stage-wise optimization settings, data scale, and compute cost so that the method itself can be implemented and evaluated with alternative components. Validation with a fully open-source teacher--student pipeline remains important future work.

\textbf{Model scaling and conditioning.}
We have not explored very large f32 backbones (e.g., $>10$B parameters), and the current model uses T5-XXL rather than a modern vision-language model for conditioning. Studying larger backbones, stronger multimodal conditioning, and additional DiT families is left for future work.

\section{Implementation Details}
\label{suppsec:implement}
\subsection{Training Strategy}
To train our model, we employ a multi-stage training strategy designed to stabilize the learning process.

\vspace{0.5em}
\noindent \textbf{Phase 1: F32 Pre-training.} We first train the DiT model in the f32 latent space for 120,000 iterations. This model is initialized from an f8 model, while the patch embedding and output prediction heads are randomly initialized. We refer to this checkpoint as \textit{f32-phase1}.

\noindent \textbf{Phase 2: Teacher Preparation.} For the teacher model training, we utilize a dual-branch initialization: the f8 branch is initialized with the f8 model, and the f32 branch is initialized with the \textit{f32-phase1} checkpoint. We train this teacher model for 5,000 iterations at resolutions of $512^2$ and $1024^2$ of varying aspect ratio.

\noindent \textbf{Phase 3: Distillation and Scaling.} Following teacher preparation, we freeze the teacher model and initialize a student model using \textit{f32-phase1} to perform distillation. The student model is first trained at resolutions of 512 and 1024 for 50,000 iterations. Subsequently, we scale the training resolution to 1K, 2K, 3K, and 4K mixed resolution for 10,000 iterations.

\paragraph{High-resolution distillation protocol.}
Our MixTeacher is prepared at 512 and 1024 resolutions, where the teacher provides stable and reliable supervision. We therefore apply the distillation loss only during the 512--1K training stage. For higher resolutions, including 2K, 3K, and 4K, we disable the teacher-based distillation loss and continue training the student using the standard flow-matching objective. This design avoids relying on an unstable or prohibitively expensive high-resolution teacher. In practice, the 1K distillation stage transfers the high-fidelity f8 priors into the f32 student, while the subsequent high-resolution flow-matching stage adapts the same student to larger resolutions efficiently.

\noindent\textbf{Loss Function:} For 1K resolution, we employ a distillation loss and flow matching loss. For higher resolutions (2K, 3K, 4K), we exclusively calculate the flow matching loss to maximize computational efficiency.

\noindent \textbf{Distributed Training.} All training is conducted on 64 NVIDIA A100 GPUs. We utilize Fully Sharded Data Parallel (FSDP) with a hybrid sharding strategy to manage memory consumption effectively during distributed training.
\subsection{Training Cost and Practical Setup}

The f8/f32 VAEs are trained separately before UltraDiff and then kept fixed, while the f8 text-to-image anchor is treated as an available pretrained component. Table~\ref{tab:training_resources} in the main paper reports the stage-wise optimization steps and aggregate GPU-days required to adapt these fixed components into the final f32 student. The multi-stage process is used only during model development; inference uses the final standalone f32 student without the teacher or f8 branch.
\subsection{Model, Data, and Open-source Plan}

The strong f8 text-to-image model used in our pipeline is an internal proprietary model. The filtered corpus size is reported in the main paper, and stage-wise training resources are summarized in Table~\ref{tab:training_resources}. All data were collected and curated under copyright-compliant data governance.

For scale context, SD3-medium reports 1B pretraining images followed by 30M aesthetic fine-tuning images and 3M preference images~\citep{esser2024scaling}. PixArt-$\Sigma$ reports a 33M-image high-resolution corpus and is initialized from PixArt-$\alpha$, whose pretraining used 25M images~\citep{chen2024pixart,chen2023pixart}. SDXL, Playground v2.5, FLUX.1, and Sana do not disclose directly comparable unique-image counts in their primary reports. Because filtering criteria, repeated sampling, and training stages differ across systems, these figures provide scale context rather than a controlled data-efficiency comparison.

We plan to release the training and inference code for UltraDiff, including the MixTeacher training pipeline and marginal score distillation procedure. However, since the current implementation relies on internal teacher/student backbones and proprietary training data, the exact pretrained weights used in this work may not be releasable.

UltraDiff is not tied to a specific proprietary implementation, but the present experiments validate only one DiT family. The framework can in principle be instantiated with open-source f8 text-to-image models and compressed-latent f32-style DiT backbones without changing the student's inference architecture. Empirical validation on fully open-source and substantially different backbones remains future work.

\section{Detailed Model Architectures and Hyperparameters}
\label{app:model_details}

In this section, we provide the specific architectural hyperparameters for the Diffusion Transformer (DiT), the Variational Autoencoders (VAEs), and the conditioning models used in UltraDiff.

\subsection{Diffusion Transformer (DiT) Architecture}
We employ a shared DiT backbone for both the teacher (dual-branch) and the student (single-branch). The model processes text embeddings (from T5-XXL) and image latents as a concatenated sequence.
Crucially, since the f32 latent space is highly compressed, we utilize a patch size $p=1$ for patch embedding layer.

Table~\ref{tab:dit_arch} summarizes the configuration of our DiT backbone.

\begin{table}[h]
    \centering
    \caption{\textbf{UltraDiff DiT Architecture Specifications.} The model follows a standard Transformer backbone with Rotary Positional Embeddings (RoPE) and QK-Norm.}
    \label{tab:dit_arch}
    \vspace{5pt}
    \begin{tabular}{l|l}
        \toprule
        \textbf{Hyperparameter} & \textbf{Value} \\
        \midrule
        Model Depth ($N$) & 28 layers \\
        Hidden Dimension ($D$) & 2048 \\
        Attention Heads & 16 \\
        KV Heads & 16  \\
        Head Dimension & $2048 / 16 = 128$ \\
        FFN Multiplier & 1.3 (SwiGLU variant) \\
        \midrule
        \textbf{Input Processing} & \\
        Input Patch Size & $1 \times 1$  \\
        Latent Channels (f8 branch) & 16 \\
        Latent Channels (f32 branch) & 64 \\
        Positional Embedding & 3D Rotary (RoPE) \\
        RoPE Base ($\theta$) & 10,000 \\
        \midrule
        \textbf{Conditioning} & \\
        Text Encoder & T5-XXL  \\
        Text Embedding Dim & 4096 \\
        \bottomrule
    \end{tabular}
\end{table}

\subsection{VAE Architectures and Training Details}
\label{app:vae_details}

We use two KL-regularized continuous VAEs. Each encoder predicts the mean and variance of a diagonal Gaussian distribution. Table~\ref{tab:vae_comparison} compares their architectures.

\paragraph{VAE reconstruction quality.}
Both VAEs are trained from scratch and then frozen during diffusion-model training. Table~\ref{tab:vae_recon} compares them with public autoencoders under the same held-out reconstruction protocol. The scale and channel count refer to each VAE's native latent output, before any model-specific patchification. FLUX.2 is therefore listed as f8/C32; its subsequent $2\times2$ patchification is part of the DiT input pipeline, not the VAE.

Our f32 VAE has lower reconstruction fidelity than our f8 VAE, consistent with its stronger information bottleneck. Results among public f32 VAEs are mixed: HunyuanImage-2.1 has higher PSNR and SSIM than our f32 VAE, but worse LPIPS, rFID, and PatchFID. Reconstruction quality is an important constraint, but it does not fully determine downstream generation quality; the diffusability of the latent space also matters~\citep{skorokhodov2025improving}.

\begin{table}[t]
\centering
\caption{\textbf{VAE reconstruction quality.} Scale and channels describe the native VAE latent output. All models are evaluated under the same reconstruction protocol.}
\label{tab:vae_recon}
\small
\setlength{\tabcolsep}{5pt}
\renewcommand{\arraystretch}{1.08}
\resizebox{0.92\linewidth}{!}{%
\begin{tabular}{lccccccc}
\toprule
\rowcolor{gray!10}
\textbf{VAE} & \textbf{Scale} & \textbf{Channels} & \textbf{PSNR}$\uparrow$ & \textbf{SSIM}$\uparrow$ & \textbf{LPIPS}$\downarrow$ & \textbf{rFID}$\downarrow$ & \textbf{PatchFID}$\downarrow$ \\
\midrule
Ours (f8) & f8 & 16 & 30.9588 & 0.8719 & 0.0343 & 2.1645 & 3.7048 \\
Ours (f32) & f32 & 64 & 26.6258 & 0.7416 & 0.0681 & 6.7395 & 3.2330 \\
SANA 1.5 DC-AE~\citep{xie2024sana} & f32 & 32 & 25.3647 & 0.6888 & 0.0930 & 7.8557 & 4.2969 \\
FLUX.2 autoencoder~\citep{flux2} & f8 & 32 & 33.8557 & 0.9242 & 0.0144 & 1.1360 & 0.9926 \\
HunyuanImage-2.1 VAE~\citep{hunyuanimage21_github} & f32 & 64 & 29.4606 & 0.8217 & 0.1111 & 9.7390 & 10.1938 \\
Stable Diffusion 3.5 VAE~\citep{esser2024scaling} & f8 & 16 & 30.6493 & 0.8609 & 0.0298 & 2.4047 & 1.3692 \\
\bottomrule
\end{tabular}}
\end{table}

All f32 variants in Table~\ref{tab:f32train} use the same f32 VAE. The GenEval gain from 0.61 with f8 initialization to 0.72 with UltraDiff therefore reflects better modeling of the same compressed latent space, rather than a change in autoencoder reconstruction quality.

\paragraph{f8 VAE.}
For the f8 branch, we utilize a standard 2D KL-VAE architecture.
\begin{itemize}
    \item \textbf{Encoder \& Decoder:} The encoder downsamples the input by a factor of $8$ spatially. It utilizes a channel multiplier strategy of $[1, 2, 4, 6]$ with a base channel width of 128.
    \item \textbf{Continuous Latent Space:} The encoder outputs a distribution parameters map with $z=16$ channels. We apply a slight KL-divergence penalty to regularize this continuous space towards a standard normal distribution.
    \item \textbf{Upsampling:} The decoder employs a \texttt{PixelShuffle} strategy for upsampling, preserving high-frequency details.
\end{itemize}

\paragraph{f32 VAE.}
The f32 VAE is a specialized deep 2D KL-VAE designed for extreme spatial compression ($32\times$).
\begin{itemize}
    \item \textbf{Encoder \& Decoder:} The encoder achieves a $32\times$ downsampling factor through a deep 6-stage network with channel multipliers $[1, 2, 4, 6, 8, 8]$. It uses standard 2D Convolutions and \texttt{GroupNorm}.
    \item \textbf{Latent Space:} To compensate for the significant loss of spatial resolution, the continuous latent dimension is increased to $z=64$.
    \item \textbf{Upsampling:} The decoder uses a \texttt{Nearest+Conv} upsampling strategy, providing robust reconstruction for the highly compressed latent grid.
\end{itemize}

\begin{table}[h]
    \centering
    \caption{\textbf{Detailed Hyperparameters for f8 and f32 VAEs.} Both models utilize 2D KL-regularized architectures with continuous latent spaces. The f32 model features a deeper network and larger latent channels to handle the $32\times$ compression rate.}
    \label{tab:vae_comparison}
    \vspace{5pt}
    \small
    \begin{tabular}{l|c|c}
        \toprule
        \textbf{Configuration} & \textbf{f8 VAE (Teacher Anchor)} & \textbf{f32 VAE (Target)} \\
        \midrule
        \textbf{Type} & 2D KL-VAE & 2D KL-VAE \\
        \textbf{Latent Space} & Continuous (Gaussian) & Continuous (Gaussian) \\
        \textbf{Downsampling Factor} & $8\times$ & $32\times$ \\
        \textbf{Latent Channels ($C$)} & 16 & 64 \\
        \midrule
        \textbf{Encoder} & & \\
        Channel Base & 128 & 128 \\
        Channel Multipliers & $[1, 2, 4, 6]$ & $[1, 2, 4, 6, 8, 8]$ \\
        ResBlocks per Level & 2 & 2 \\
        \midrule
        \textbf{Decoder} & & \\
        Channel Multipliers & $[6, 4, 2, 1]$ & $[8, 8, 6, 4, 2, 1]$ \\
        Upsampling Method & Residual PixelShuffle & Nearest + Conv \\
        \midrule
        \textbf{Loss weighting} & & \\
        Perceptual Loss (LPIPS) & 1.0 & 0.5 \\
        KL Weight & $1 \times 10^{-12}$ & $1 \times 10^{-12}$ \\
        Loss Formulation & LPVAE & LPVAE \\
        \bottomrule
    \end{tabular}
\end{table}

\subsection{Latent Normalization}
A critical component of our method is addressing the distribution shift between the two continuous latent spaces. We apply channel-wise normalization before feeding latents into the DiT.

\subsection{Training Hyperparameters}
We train the models using the AdamW optimizer with a linear warmup and cosine decay scheduler. The specific training settings are detailed in Table~\ref{tab:train_params}.

\begin{table}[h]
    \centering
    \caption{\textbf{Training hyperparameters.} Common optimizer settings and stage-specific training lengths used in the UltraDiff pipeline.}
    \label{tab:train_params}
    \vspace{5pt}
    \begin{tabular}{l|l}
        \toprule
        \textbf{Parameter} & \textbf{Value} \\
        \midrule
        Optimizer & AdamW \\
        Adam $\beta_1, \beta_2$ & 0.9, 0.95 \\
        Learning Rate (LR) & $1 \times 10^{-4}$ \\
        Weight Decay & 0.01 \\
        LR Schedule & Linear Warmup + Cosine Decay \\
        Warmup Steps & 500 \\
        Min LR & $1 \times 10^{-5}$ \\
        \midrule
        Gradient Clipping & 1.0 \\
        EMA Decay & 0.9999 \\
        EMA Start Step & 2,000 \\
        Batch Size & 2048 (Global) \\
        Training Steps & Phase 1: 120K; Phase 2: 5K; Phase 3: 50K + 10K \\
        \midrule
        \textbf{Loss Weighting} & \\
        Prediction Target & Velocity ($v$-prediction) \\
        Loss Type & MSE ($L_2$) \\
        Noise Scheduler & Linear Interpolant Scheduler \\
        \bottomrule
    \end{tabular}
\end{table}

\newpage
\section*{NeurIPS Paper Checklist}

The checklist is designed to encourage best practices for responsible machine learning research, addressing issues of reproducibility, transparency, research ethics, and societal impact. Do not remove the checklist: {\bf The papers not including the checklist will be desk rejected.} The checklist should follow the references and follow the (optional) supplemental material.  The checklist does NOT count towards the page
limit. 

Please read the checklist guidelines carefully for information on how to answer these questions. For each question in the checklist:
\begin{itemize}
    \item You should answer \answerYes{}, \answerNo{}, or \answerNA{}.
    \item \answerNA{} means either that the question is Not Applicable for that particular paper or the relevant information is Not Available.
    \item Please provide a short (1--2 sentence) justification right after your answer (even for \answerNA). 
\end{itemize}

{\bf The checklist answers are an integral part of your paper submission.} They are visible to the reviewers, area chairs, senior area chairs, and ethics reviewers. You will also be asked to include it (after eventual revisions) with the final version of your paper, and its final version will be published with the paper.

The reviewers of your paper will be asked to use the checklist as one of the factors in their evaluation. While \answerYes{} is generally preferable to \answerNo{}, it is perfectly acceptable to answer \answerNo{} provided a proper justification is given (e.g., error bars are not reported because it would be too computationally expensive'' or ``we were unable to find the license for the dataset we used''). In general, answering \answerNo{} or \answerNA{} is not grounds for rejection. While the questions are phrased in a binary way, we acknowledge that the true answer is often more nuanced, so please just use your best judgment and write a justification to elaborate. All supporting evidence can appear either in the main paper or the supplemental material, provided in appendix. If you answer \answerYes{} to a question, in the justification please point to the section(s) where related material for the question can be found.

IMPORTANT, please:
\begin{itemize}
    \item {\bf Delete this instruction block, but keep the section heading ``NeurIPS Paper Checklist"},
    \item  {\bf Keep the checklist subsection headings, questions/answers and guidelines below.}
    \item {\bf Do not modify the questions and only use the provided macros for your answers}.
\end{itemize}


\begin{enumerate}

\item {\bf Claims}
    \item[] Question: Do the main claims made in the abstract and introduction accurately reflect the paper's contributions and scope?
    \item[] Answer: \answerYes{} 
    \item[] Justification: See abstract and introduction.
    \item[] Guidelines:
    \begin{itemize}
        \item The answer \answerNA{} means that the abstract and introduction do not include the claims made in the paper.
        \item The abstract and/or introduction should clearly state the claims made, including the contributions made in the paper and important assumptions and limitations. A \answerNo{} or \answerNA{} answer to this question will not be perceived well by the reviewers. 
        \item The claims made should match theoretical and experimental results, and reflect how much the results can be expected to generalize to other settings. 
        \item It is fine to include aspirational goals as motivation as long as it is clear that these goals are not attained by the paper. 
    \end{itemize}

\item {\bf Limitations}
    \item[] Question: Does the paper discuss the limitations of the work performed by the authors?
    \item[] Answer: \answerYes{} 
    \item[] Justification: See Section~\ref{app:limitations}
    \item[] Guidelines:
    \begin{itemize}
        \item The answer \answerNA{} means that the paper has no limitation while the answer \answerNo{} means that the paper has limitations, but those are not discussed in the paper. 
        \item The authors are encouraged to create a separate ``Limitations'' section in their paper.
        \item The paper should point out any strong assumptions and how robust the results are to violations of these assumptions (e.g., independence assumptions, noiseless settings, model well-specification, asymptotic approximations only holding locally). The authors should reflect on how these assumptions might be violated in practice and what the implications would be.
        \item The authors should reflect on the scope of the claims made, e.g., if the approach was only tested on a few datasets or with a few runs. In general, empirical results often depend on implicit assumptions, which should be articulated.
        \item The authors should reflect on the factors that influence the performance of the approach. For example, a facial recognition algorithm may perform poorly when image resolution is low or images are taken in low lighting. Or a speech-to-text system might not be used reliably to provide closed captions for online lectures because it fails to handle technical jargon.
        \item The authors should discuss the computational efficiency of the proposed algorithms and how they scale with dataset size.
        \item If applicable, the authors should discuss possible limitations of their approach to address problems of privacy and fairness.
        \item While the authors might fear that complete honesty about limitations might be used by reviewers as grounds for rejection, a worse outcome might be that reviewers discover limitations that aren't acknowledged in the paper. The authors should use their best judgment and recognize that individual actions in favor of transparency play an important role in developing norms that preserve the integrity of the community. Reviewers will be specifically instructed to not penalize honesty concerning limitations.
    \end{itemize}

\item {\bf Theory assumptions and proofs}
    \item[] Question: For each theoretical result, does the paper provide the full set of assumptions and a complete (and correct) proof?
    \item[] Answer: \answerNA{} 
    \item[] Justification: \answerNA{}
    \item[] Guidelines:
    \begin{itemize}
        \item The answer \answerNA{} means that the paper does not include theoretical results. 
        \item All the theorems, formulas, and proofs in the paper should be numbered and cross-referenced.
        \item All assumptions should be clearly stated or referenced in the statement of any theorems.
        \item The proofs can either appear in the main paper or the supplemental material, but if they appear in the supplemental material, the authors are encouraged to provide a short proof sketch to provide intuition. 
        \item Inversely, any informal proof provided in the core of the paper should be complemented by formal proofs provided in appendix or supplemental material.
        \item Theorems and Lemmas that the proof relies upon should be properly referenced. 
    \end{itemize}

    \item {\bf Experimental result reproducibility}
    \item[] Question: Does the paper fully disclose all the information needed to reproduce the main experimental results of the paper to the extent that it affects the main claims and/or conclusions of the paper (regardless of whether the code and data are provided or not)?
    \item[] Answer: \answerYes{} 
    \item[] Justification: See Section~\ref{sec:experiment} and appendix
    \item[] Guidelines:
    \begin{itemize}
        \item The answer \answerNA{} means that the paper does not include experiments.
        \item If the paper includes experiments, a \answerNo{} answer to this question will not be perceived well by the reviewers: Making the paper reproducible is important, regardless of whether the code and data are provided or not.
        \item If the contribution is a dataset and\slash or model, the authors should describe the steps taken to make their results reproducible or verifiable. 
        \item Depending on the contribution, reproducibility can be accomplished in various ways. For example, if the contribution is a novel architecture, describing the architecture fully might suffice, or if the contribution is a specific model and empirical evaluation, it may be necessary to either make it possible for others to replicate the model with the same dataset, or provide access to the model. In general. releasing code and data is often one good way to accomplish this, but reproducibility can also be provided via detailed instructions for how to replicate the results, access to a hosted model (e.g., in the case of a large language model), releasing of a model checkpoint, or other means that are appropriate to the research performed.
        \item While NeurIPS does not require releasing code, the conference does require all submissions to provide some reasonable avenue for reproducibility, which may depend on the nature of the contribution. For example
        \begin{enumerate}
            \item If the contribution is primarily a new algorithm, the paper should make it clear how to reproduce that algorithm.
            \item If the contribution is primarily a new model architecture, the paper should describe the architecture clearly and fully.
            \item If the contribution is a new model (e.g., a large language model), then there should either be a way to access this model for reproducing the results or a way to reproduce the model (e.g., with an open-source dataset or instructions for how to construct the dataset).
            \item We recognize that reproducibility may be tricky in some cases, in which case authors are welcome to describe the particular way they provide for reproducibility. In the case of closed-source models, it may be that access to the model is limited in some way (e.g., to registered users), but it should be possible for other researchers to have some path to reproducing or verifying the results.
        \end{enumerate}
    \end{itemize}

\item {\bf Open access to data and code}
    \item[] Question: Does the paper provide open access to the data and code, with sufficient instructions to faithfully reproduce the main experimental results, as described in supplemental material?
    \item[] Answer: \answerNA{} 
    \item[] Justification: We will opensource upon acceptance of this manuscript.
    \item[] Guidelines:
    \begin{itemize}
        \item The answer \answerNA{} means that paper does not include experiments requiring code.
        \item Please see the NeurIPS code and data submission guidelines (\url{https://neurips.cc/public/guides/CodeSubmissionPolicy}) for more details.
        \item While we encourage the release of code and data, we understand that this might not be possible, so \answerNo{} is an acceptable answer. Papers cannot be rejected simply for not including code, unless this is central to the contribution (e.g., for a new open-source benchmark).
        \item The instructions should contain the exact command and environment needed to run to reproduce the results. See the NeurIPS code and data submission guidelines (\url{https://neurips.cc/public/guides/CodeSubmissionPolicy}) for more details.
        \item The authors should provide instructions on data access and preparation, including how to access the raw data, preprocessed data, intermediate data, and generated data, etc.
        \item The authors should provide scripts to reproduce all experimental results for the new proposed method and baselines. If only a subset of experiments are reproducible, they should state which ones are omitted from the script and why.
        \item At submission time, to preserve anonymity, the authors should release anonymized versions (if applicable).
        \item Providing as much information as possible in supplemental material (appended to the paper) is recommended, but including URLs to data and code is permitted.
    \end{itemize}

\item {\bf Experimental setting/details}
    \item[] Question: Does the paper specify all the training and test details (e.g., data splits, hyperparameters, how they were chosen, type of optimizer) necessary to understand the results?
    \item[] Answer: \answerYes{} 
    \item[] Justification: See experiment.
    \item[] Guidelines:
    \begin{itemize}
        \item The answer \answerNA{} means that the paper does not include experiments.
        \item The experimental setting should be presented in the core of the paper to a level of detail that is necessary to appreciate the results and make sense of them.
        \item The full details can be provided either with the code, in appendix, or as supplemental material.
    \end{itemize}

\item {\bf Experiment statistical significance}
    \item[] Question: Does the paper report error bars suitably and correctly defined or other appropriate information about the statistical significance of the experiments?
    \item[] Answer: \answerYes{} 
    \item[] Justification: See experiment.
    \item[] Guidelines:
    \begin{itemize}
        \item The answer \answerNA{} means that the paper does not include experiments.
        \item The authors should answer \answerYes{} if the results are accompanied by error bars, confidence intervals, or statistical significance tests, at least for the experiments that support the main claims of the paper.
        \item The factors of variability that the error bars are capturing should be clearly stated (for example, train/test split, initialization, random drawing of some parameter, or overall run with given experimental conditions).
        \item The method for calculating the error bars should be explained (closed form formula, call to a library function, bootstrap, etc.)
        \item The assumptions made should be given (e.g., Normally distributed errors).
        \item It should be clear whether the error bar is the standard deviation or the standard error of the mean.
        \item It is OK to report 1-sigma error bars, but one should state it. The authors should preferably report a 2-sigma error bar than state that they have a 96\% CI, if the hypothesis of Normality of errors is not verified.
        \item For asymmetric distributions, the authors should be careful not to show in tables or figures symmetric error bars that would yield results that are out of range (e.g., negative error rates).
        \item If error bars are reported in tables or plots, the authors should explain in the text how they were calculated and reference the corresponding figures or tables in the text.
    \end{itemize}

\item {\bf Experiments compute resources}
    \item[] Question: For each experiment, does the paper provide sufficient information on the computer resources (type of compute workers, memory, time of execution) needed to reproduce the experiments?
    \item[] Answer: \answerYes{} 
    \item[] Justification: See experiment
    \item[] Guidelines:
    \begin{itemize}
        \item The answer \answerNA{} means that the paper does not include experiments.
        \item The paper should indicate the type of compute workers CPU or GPU, internal cluster, or cloud provider, including relevant memory and storage.
        \item The paper should provide the amount of compute required for each of the individual experimental runs as well as estimate the total compute. 
        \item The paper should disclose whether the full research project required more compute than the experiments reported in the paper (e.g., preliminary or failed experiments that didn't make it into the paper). 
    \end{itemize}
    
\item {\bf Code of ethics}
    \item[] Question: Does the research conducted in the paper conform, in every respect, with the NeurIPS Code of Ethics \url{https://neurips.cc/public/EthicsGuidelines}?
    \item[] Answer: \answerYes{} 
    \item[] Justification: \answerNA{}
    \item[] Guidelines:
    \begin{itemize}
        \item The answer \answerNA{} means that the authors have not reviewed the NeurIPS Code of Ethics.
        \item If the authors answer \answerNo, they should explain the special circumstances that require a deviation from the Code of Ethics.
        \item The authors should make sure to preserve anonymity (e.g., if there is a special consideration due to laws or regulations in their jurisdiction).
    \end{itemize}

\item {\bf Broader impacts}
    \item[] Question: Does the paper discuss both potential positive societal impacts and negative societal impacts of the work performed?
    \item[] Answer: \answerNA{} 
    \item[] Justification: See Broader impacts
    \item[] Guidelines:
    \begin{itemize}
        \item The answer \answerNA{} means that there is no societal impact of the work performed.
        \item If the authors answer \answerNA{} or \answerNo, they should explain why their work has no societal impact or why the paper does not address societal impact.
        \item Examples of negative societal impacts include potential malicious or unintended uses (e.g., disinformation, generating fake profiles, surveillance), fairness considerations (e.g., deployment of technologies that could make decisions that unfairly impact specific groups), privacy considerations, and security considerations.
        \item The conference expects that many papers will be foundational research and not tied to particular applications, let alone deployments. However, if there is a direct path to any negative applications, the authors should point it out. For example, it is legitimate to point out that an improvement in the quality of generative models could be used to generate Deepfakes for disinformation. On the other hand, it is not needed to point out that a generic algorithm for optimizing neural networks could enable people to train models that generate Deepfakes faster.
        \item The authors should consider possible harms that could arise when the technology is being used as intended and functioning correctly, harms that could arise when the technology is being used as intended but gives incorrect results, and harms following from (intentional or unintentional) misuse of the technology.
        \item If there are negative societal impacts, the authors could also discuss possible mitigation strategies (e.g., gated release of models, providing defenses in addition to attacks, mechanisms for monitoring misuse, mechanisms to monitor how a system learns from feedback over time, improving the efficiency and accessibility of ML).
    \end{itemize}
    
\item {\bf Safeguards}
    \item[] Question: Does the paper describe safeguards that have been put in place for responsible release of data or models that have a high risk for misuse (e.g., pre-trained language models, image generators, or scraped datasets)?
    \item[] Answer: \answerNo 
    \item[] Justification: \answerNo
    \item[] Guidelines:
    \begin{itemize}
        \item The answer \answerNA{} means that the paper poses no such risks.
        \item Released models that have a high risk for misuse or dual-use should be released with necessary safeguards to allow for controlled use of the model, for example by requiring that users adhere to usage guidelines or restrictions to access the model or implementing safety filters. 
        \item Datasets that have been scraped from the Internet could pose safety risks. The authors should describe how they avoided releasing unsafe images.
        \item We recognize that providing effective safeguards is challenging, and many papers do not require this, but we encourage authors to take this into account and make a best faith effort.
    \end{itemize}

\item {\bf Licenses for existing assets}
    \item[] Question: Are the creators or original owners of assets (e.g., code, data, models), used in the paper, properly credited and are the license and terms of use explicitly mentioned and properly respected?
    \item[] Answer: \answerYes{} 
    \item[] Justification: \answerNo
    \item[] Guidelines:
    \begin{itemize}
        \item The answer \answerNA{} means that the paper does not use existing assets.
        \item The authors should cite the original paper that produced the code package or dataset.
        \item The authors should state which version of the asset is used and, if possible, include a URL.
        \item The name of the license (e.g., CC-BY 4.0) should be included for each asset.
        \item For scraped data from a particular source (e.g., website), the copyright and terms of service of that source should be provided.
        \item If assets are released, the license, copyright information, and terms of use in the package should be provided. For popular datasets, \url{paperswithcode.com/datasets} has curated licenses for some datasets. Their licensing guide can help determine the license of a dataset.
        \item For existing datasets that are re-packaged, both the original license and the license of the derived asset (if it has changed) should be provided.
        \item If this information is not available online, the authors are encouraged to reach out to the asset's creators.
    \end{itemize}

\item {\bf New assets}
    \item[] Question: Are new assets introduced in the paper well documented and is the documentation provided alongside the assets?
    \item[] Answer: \answerNo 
    \item[] Justification: \answerNo
    \item[] Guidelines:
    \begin{itemize}
        \item The answer \answerNA{} means that the paper does not release new assets.
        \item Researchers should communicate the details of the dataset\slash code\slash model as part of their submissions via structured templates. This includes details about training, license, limitations, etc. 
        \item The paper should discuss whether and how consent was obtained from people whose asset is used.
        \item At submission time, remember to anonymize your assets (if applicable). You can either create an anonymized URL or include an anonymized zip file.
    \end{itemize}

\item {\bf Crowdsourcing and research with human subjects}
    \item[] Question: For crowdsourcing experiments and research with human subjects, does the paper include the full text of instructions given to participants and screenshots, if applicable, as well as details about compensation (if any)? 
    \item[] Answer: \answerNA{} 
    \item[] Justification: \answerNA{}
    \item[] Guidelines:
    \begin{itemize}
        \item The answer \answerNA{} means that the paper does not involve crowdsourcing nor research with human subjects.
        \item Including this information in the supplemental material is fine, but if the main contribution of the paper involves human subjects, then as much detail as possible should be included in the main paper. 
        \item According to the NeurIPS Code of Ethics, workers involved in data collection, curation, or other labor should be paid at least the minimum wage in the country of the data collector. 
    \end{itemize}

\item {\bf Institutional review board (IRB) approvals or equivalent for research with human subjects}
    \item[] Question: Does the paper describe potential risks incurred by study participants, whether such risks were disclosed to the subjects, and whether Institutional Review Board (IRB) approvals (or an equivalent approval/review based on the requirements of your country or institution) were obtained?
    \item[] Answer: \answerNA{} 
    \item[] Justification: 
    \item[] Guidelines:
    \begin{itemize}
        \item The answer \answerNA{} means that the paper does not involve crowdsourcing nor research with human subjects.
        \item Depending on the country in which research is conducted, IRB approval (or equivalent) may be required for any human subjects research. If you obtained IRB approval, you should clearly state this in the paper. 
        \item We recognize that the procedures for this may vary significantly between institutions and locations, and we expect authors to adhere to the NeurIPS Code of Ethics and the guidelines for their institution. 
        \item For initial submissions, do not include any information that would break anonymity (if applicable), such as the institution conducting the review.
    \end{itemize}

\item {\bf Declaration of LLM usage}
    \item[] Question: Does the paper describe the usage of LLMs if it is an important, original, or non-standard component of the core methods in this research? Note that if the LLM is used only for writing, editing, or formatting purposes and does \emph{not} impact the core methodology, scientific rigor, or originality of the research, declaration is not required.
    \item[] Answer: \answerNA{} 
    \item[] Justification: 
    \item[] Guidelines:
    \begin{itemize}
        \item The answer \answerNA{} means that the core method development in this research does not involve LLMs as any important, original, or non-standard components.
        \item Please refer to our LLM policy in the NeurIPS handbook for what should or should not be described.
    \end{itemize}

\end{enumerate}

\end{document}